\documentclass[11pt,twoside]{cis-tech}
\usepackage{epsfig,rotating,fancybox,amsmath,alltt,boxedminipage}
\title{A Formal Framework for Linguistic Annotation}
\author{Steven Bird and Mark Liberman \\[2ex]
Linguistic Data Consortium, University of Pennsylvania\\
3615 Market St, Philadelphia, PA 19104-2608, USA\\
Email: {\tt \{sb,myl\}@ldc.upenn.edu}}
\date{Technical Report MS-CIS-99-01\\
Department of Computer and Information Science\\[2ex]
March 1999}

\newenvironment{pagefig}[1]%
  {\begin{figure}[p]\begin{sideways}\begin{minipage}{\textheight}\vspace*{#1}}%
  {\end{minipage}\end{sideways}\end{figure}}
\newenvironment{bv}{\scriptsize\hspace*{.25in}\begin{boxedminipage}[t]{6in}%
  \begin{alltt}}{\end{alltt}\end{boxedminipage}\normalsize\vspace*{1ex}}
\newenvironment{sv}{\scriptsize\begin{alltt}}{\end{alltt}\normalsize}

\usepackage{fancyheadings}
\def\runninghead{A Formal Framework for Linguistic Annotation}
\def\type{\operatorname{type}}
\def\glb{\operatorname{glb}}
\def\lub{\operatorname{lub}}
\def\dom{\operatorname{dom}}

\def\au#1#2{\ar@/^/[#1]^{#2}}
\def\ad#1#2{\ar@/_/[#1]_{#2}}
\def\larc#1{\ar@/^/[r]^{#1}}
\def\arc{\ar@/^/[r]}
\def\smtt#1{{\small\tt #1}}

\begin{document}
\maketitle\thispagestyle{tech}\vfil

\begin{abstract}{\normalsize
`Linguistic annotation' covers any descriptive or analytic 
notations applied to raw language data. The basic data may be in the form of 
time functions -- audio, video and/or physiological recordings -- or it may 
be textual. The added notations may include transcriptions of all sorts (from 
phonetic features to discourse structures), part-of-speech and sense tagging, 
syntactic analysis, `named entity' identification, co-reference annotation, 
and so on.
While there are several ongoing efforts to provide formats
and tools for such annotations and to publish annotated linguistic
databases, the lack of widely accepted standards is becoming a critical
problem.  Proposed standards, to the extent they exist, have focussed
on file formats.  This paper focuses instead on the logical
structure of linguistic annotations.  We survey a wide variety of
existing annotation formats and demonstrate a common conceptual core,
the {\em annotation graph}.  This
provides a formal framework for constructing, maintaining and searching
linguistic annotations,
while remaining consistent with many alternative data structures
and file formats.
}\end{abstract}
\vfil\pagebreak

\section{Introduction}

In the simplest and commonest case, `linguistic annotation' is an
orthographic transcription of speech, time-aligned to an audio or
video recording.  Other central examples include morphological
analysis, part-of-speech tagging and syntactic bracketing; phonetic
segmentation and labeling; annotation of disfluencies, prosodic
phrasing, intonation, gesture, and discourse structure; marking of
co-reference, `named entity' tagging, and sense tagging; and
phrase-level or word-level translations.  Linguistic annotations may
describe texts or recorded signals. Our focus will be on the latter,
broadly construed to include any kind of audio, video or physiological
recording, or any combination of these, for which we will use the
cover term `linguistic signals'. However, our ideas also apply to the
annotation of texts.


Linguistic annotations have seen increasingly broad use in the scientific study
of language, in research and development of language-related
technologies, and in language-related applications more broadly, for
instance in the entertainment industry. Particular cases range from
speech databases used in speech recognition or speech synthesis
development, to annotated ethnographic materials, to cartoon sound
tracks. There have been many independent efforts to provide tools for
creating linguistic annotations, to provide general formats
for expressing them, and to provide tools for creating, browsing and
searching databases containing them -- see [\smtt{www.ldc.upenn.edu/annotation}].
Within the area of speech and
language technology development alone, hundreds of annotated
linguistic databases have been published in the past fifteen years.


While the utility of existing tools, formats and databases is
unquestionable, their sheer variety -- and the lack of standards able
to mediate among them -- is becoming a critical problem.
Particular bodies of data are
created with particular needs in mind, using formats and tools
tailored to those needs, based on the resources and practices of the
community involved.  Once created, a linguistic database
may subsequently be used for a variety of unforeseen purposes,
both inside and outside the community that created it.  Adapting
existing software for creation, update, indexing, search and display
of `foreign' databases typically requires extensive
re-engineering. Working across a set of databases requires repeated
adaptations of this kind.

Previous attempts to standardize practice in this area have primarily
focussed on file formats and on the tags, attributes and values for
describing content (e.g.\ \cite{MacWhinney95}, \cite{UTF98}; but see
also \cite{Schiel98}).  We contend that file formats and content
specifications, though important, are secondary.  Instead, we focus on
the logical structure of linguistic annotations.  We demonstrate that,
while different existing annotations vary greatly in their form, their
logical structure is remarkably consistent.  In order to help us think
about the form and meaning of annotations, we describe a simple
mathematical framework endowed with a practically useful formal
structure.  This opens up an interesting range of new possibilities
for creation, maintenance and search.  We claim that essentially all
existing annotations can be expressed in this framework. Thus, the
framework should
provide a useful `interlingua' for translation among the multiplicity
of current annotation formats, and also should permit the development
of new tools with broad applicability.

Before we embark on our survey, a terminological aside is necessary.
As far as we are aware, there is no existing cover term for the kinds
of transcription, description and analysis that we address here.
`Transcription' may refer to the use of ordinary orthography, or a
phonetic orthography; it can plausibly be extended to certain aspects
of prosody (`intonational transcription'), but not to other kinds of
analysis (morphological, syntactic, rhetorical or discourse
structural, semantic, etc).  One does not talk about a `syntactic
transcription', although this is at least as determinate a
representation of the speech stream as is a phonetic transcription.
`Coding' has been used by social scientists to mean something like
`the assignment of events to stipulated symbolic categories,' as a
generalization of the ordinary language meaning associated with
translating words and phrases into references to a shared, secret code
book.  It would be idiosyncratic and confusing (though conceptually
plausible) to refer to ordinary orthographic transcription in this
way.  The term `markup' has come to have a specific technical meaning,
involving the addition of typographical or structural information to a
document.

In ordinary language, `annotation' means a sort of commentary or
explanation (typically indexed to particular portions of a text), or
the act of producing such a commentary.  Like `markup', this term's
ordinary meaning plausibly covers the non-transcriptional kinds of
linguistic analysis, such as the annotation of syntactic structure or
of co-reference. Some speech and language engineers have begun to use
`annotation' in this way, but there is not yet a specific,
widely-accepted technical meaning. We feel that it is reasonable to
generalize this term to cover the case of transcribing speech, by
thinking of `annotation' as the provision of any symbolic description
of particular portions of a pre-existing linguistic object.
If the object is a speech recording, then an ordinary orthographic
transcription is certainly a kind of annotation in this sense --
though it is one in which the amount of critical judgment is
small.

In sum, `annotation' is a reasonable candidate for adoption as the
needed cover term.  The alternative would be to create a neologism
(`scription'?).  Extension of the existing term `annotation' seems
preferable to us.

\section{Existing Annotation Systems}\label{sec:survey}

In order to justify our claim that essentially all existing linguistic
annotations can be expressed in the framework that we propose, we need
to discuss a representative set of such annotations. In addition,
it will be easiest to understand our proposal if we motivate it,
piece by piece, in terms of the logical structures underlying
existing annotation practice.
	
This section reviews nine bodies of annotation practice, with
a concrete example of each.
For each example, we show how to express its various structuring
conventions in terms of our `annotation graphs', which are
networks consisting of nodes and arcs, decorated with
time marks and labels.  Following the
review, we shall discuss some general architectural issues (\S\ref{sec:arch}),
give a formal presentation of the `annotation graph'
concept (\S\ref{sec:algebra}), and describe some indexing methods
(\S\ref{sec:indexing}).  The paper concludes in \S\ref{sec:conclusion}
with an evaluation
of the proposed formalism and a discussion of future work.

The nine annotation models to be discussed in detail are
TIMIT \cite{TIMIT86},
Partitur \cite{Schiel98},
CHILDES \cite{MacWhinney95},
the LACITO Archiving Project \cite{Michailovsky98},
LDC Broadcast News,
LDC Telephone Speech,
NIST UTF \cite{UTF98},
Emu \cite{Cassidy96} and
Festival \cite{Taylor98}.
These models are widely divergent in type and purpose.
Some, like TIMIT, are associated with a specific database,
others, like UTF, are associated with a specific linguistic
domain (here conversation),
while still others, like Festival, are associated with a
specific application domain (here, speech synthesis).

Several other systems and formats have been considered in developing our
ideas, but will not be discussed in detail. These include
Switchboard \cite{Godfrey92},
HCRC MapTask \cite{Anderson91},
TEI \cite{TEI-P3}, and MATE \cite{MATE-D1.2}.
The Switchboard and MapTask formats are conversational transcription
systems that encode a subset of the information in the LDC and NIST
formats cited above. The TEI guidelines for `Transcriptions of
Speech' \cite[p11]{TEI-P3} are also similar in content, though they
offer access to a very broad range of representational techniques
drawn from other aspects of the TEI specification. The TEI report
sketches or alludes to a correspondingly wide range of possible issues
in speech annotation.  All of these seem to be encompassed within our
proposed framework, but it does not seem appropriate to speculate at
much greater length about this, given that this portion of the TEI
guidelines does not seem to have been used in any published
transcriptions to date.  As for MATE, it is a new SGML- and TEI-based
standard for dialogue annotation, in the process of being
developed. It also appears to fall within the class of annotation
systems that our framework covers, but it would be premature to discuss the
correspondences in detail.
Still other models that we are aware of include
\cite{Altosaar98,Hertz90,Schegloff98}.

Note that there are many kinds of linguistic database that are not
linguistic annotations in our sense, although they may be connected
with linguistic annotations in various ways.  One example is a lexical
database with pointers to speech recordings along with transcriptions
of those recordings (e.g.\ HyperLex \cite{Bird97sigphon}).
Another example would be collections of
information that are not specific to any particular stretch of speech,
such as demographic information about speakers.
We return to such cases in \S\ref{sec:extensions}.

\subsection{TIMIT}\label{sec:timit}

The TIMIT corpus of read speech
was designed to provide data for the acquisition of
acoustic-phonetic knowledge and to support the development
and evaluation of automatic speech recognition systems.
TIMIT was the first annotated speech database to be published,
and it has been widely used and also republished in several
different forms.  It is also especially simple and clear
in structure.
Here, we just give one example taken from the TIMIT database
\cite{TIMIT86}.  The file \smtt{train/dr1/fjsp0/sa1.wrd} contains:

\begin{sv}
2360 5200 she
5200 9680 had
9680 11077 your
11077 16626 dark
16626 22179 suit
22179 24400 in
24400 30161 greasy
30161 36150 wash
36720 41839 water
41839 44680 all
44680 49066 year
\end{sv}

This file combines an ordinary string of orthographic words with
information about the starting and ending time of each word,
measured in audio samples at a sampling rate of 16 kHz.
The path name \smtt{train/dr1/fjsp0/sa1.wrd} tells us that
this is training data, from `dialect region 1', from female speaker
`jsp0', containing words and audio sample numbers.
The file \smtt{train/dr1/fjsp0/sa1.phn} contains a corresponding
broad phonetic transcription, which begins as follows:

\begin{sv}
0 2360 h#
2360 3720 sh
3720 5200 iy
5200 6160 hv
6160 8720 ae
8720 9680 dcl
9680 10173 y
10173 11077 axr
11077 12019 dcl
12019 12257 d
\end{sv}

\begin{pagefig}{20ex}
\centerline{\epsfig{figure=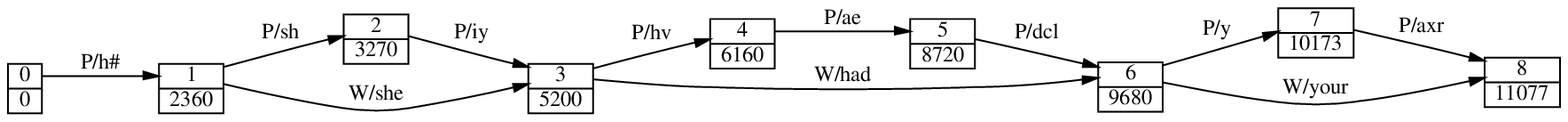,width=\linewidth}}
\caption{Graph Structure for TIMIT Example}\label{timit}
\vspace*{20ex}
\centerline{\epsfig{figure=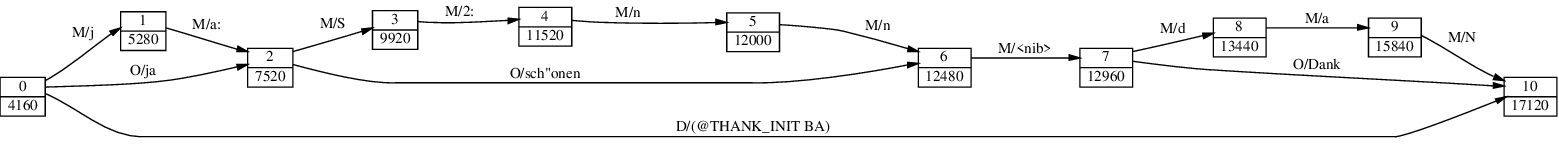,width=\linewidth}}
\caption{Graph Structure for Partitur Example}\label{partitur}
\end{pagefig}

We can interpret each line: \smtt{$<$time1$>$ $<$time2$>$ $<$label$>$}
as an edge in a directed acyclic graph, where
the two times are attributes of nodes
and the label is a property of an edge connecting those nodes.
The resulting annotation graph for the above fragment is
shown in Figure~\ref{timit}.  Observe that edge labels
have the form \smtt{$<$type$>$/$<$content$>$} where
the \smtt{$<$type$>$} here tells us what kind of label it is.
We have used \smtt{P} for the (phonetic transcription) contents of the .phn
file, and \smtt{W} for the (orthographic word) contents of the .wrd file.
The top number for each node is an arbitrary node identifier,
while the bottom number is the time reference.
We distinguish node identifiers from time references since
nodes may lack time references, as we shall see later.

\subsection{Partitur}\label{sec:partitur}

The Partitur format of the Bavarian Archive for Speech Signals
\cite{Schiel98} is founded on the collective experience of a broad
range of German speech database efforts.  The aim has been to create
`an open (that is extensible), robust format to represent results from
many different research labs in a common source.'  Partitur is
valuable because it represents a careful attempt to present a common
low-level core for all of those independent efforts, similar in spirit
to our effort here.  In essence, Partitur extends and reconceptualizes the TIMIT
format to encompass a wide range of annotation types.

The Partitur format permits time-aligned, multi-tier description of speech
signals, along with links between units on different tiers
which are independent of the temporal structure.
For ease of presentation, the example Partitur file will be broken
into a number of chunks, and certain details (such as the header)
will be ignored.  The fragment under discussion is from one
of the Verbmobil corpora at the Bavarian Archive of Speech Signals.
The KAN tier provides the
canonical transcription, and introduces a numerical identifier
for each word to serve as an anchor for all other material.

\begin{sv}
KAN: 0 j'a:
KAN: 1 S'2:n@n
KAN: 2 d'aNk
KAN: 3 das+
KAN: 4 vE:r@+
KAN: 5 z'e:6
KAN: 6 n'Et
\end{sv}

Tiers for orthographic and transliteration information
then reference these anchors as shown below, with orthographic
information (ORT) on the left and transliteration information (TRL)
on the right.

\begin{sv}
ORT: 0 ja                             TRL: 0 <A>
ORT: 1 sch"onen                       TRL: 0 ja ,
ORT: 2 Dank                           TRL: 1 sch"onen
ORT: 3 das                            TRL: 1 <:<#Klopfen>
ORT: 4 w"are                          TRL: 2 Dank:> ,
ORT: 5 sehr                           TRL: 3 das
ORT: 6 nett                           TRL: 4 w"ar'
                                      TRL: 5 sehr
                                      TRL: 6 nett . 
\end{sv}

Higher level structure representing dialogue acts
refers to extended intervals using
contiguous sequences of anchors, as shown below:

\begin{sv}
DAS: 0,1,2 @(THANK_INIT BA)
DAS: 3,4,5,6 @(FEEDBACK_ACKNOWLEDGEMENT BA)
\end{sv}

Speech data can be referenced using annotation lines containing
offset and duration information.  As before, links to the KAN
anchors are also specified (as the second-last field).

\begin{sv}
MAU: 4160 1119 0 j                    MAU: 17760 1119 3 a
MAU: 5280 2239 0 a:                   MAU: 18880 1279 3 s
MAU: 7520 2399 1 S                    MAU: 20160 959 4 v
MAU: 9920 1599 1 2:                   MAU: 21120 639 4 E:
MAU: 11520 479 1 n                    MAU: 21760 1119 4 6
MAU: 12000 479 1 n                    MAU: 22880 1119 5 z
MAU: 12480 479 -1 <nib>               MAU: 24000 799 5 e:
MAU: 12960 479 2 d                    MAU: 24800 1119 5 6
MAU: 13440 2399 2 a                   MAU: 25920 1279 6 n
MAU: 15840 1279 2 N                   MAU: 27200 1919 6 E
MAU: 17120 639 3 d                    MAU: 29120 2879 6 t
                                      MAU: 32000 2559 -1 <p:>
\end{sv}

The content of the first few words of the
ORT (orthography), DAS (dialog act) and MAU
(phonetic segment) tiers can
apparently be expressed as in Figure~\ref{partitur}.
Note that we abbreviate the types, using
\smtt{O/} for ORT, \smtt{D/} for DAS, and \smtt{M/} for MAU.

\subsection{CHILDES}
\label{sec:childes}

\begin{pagefig}{20ex}
\centerline{\epsfig{figure=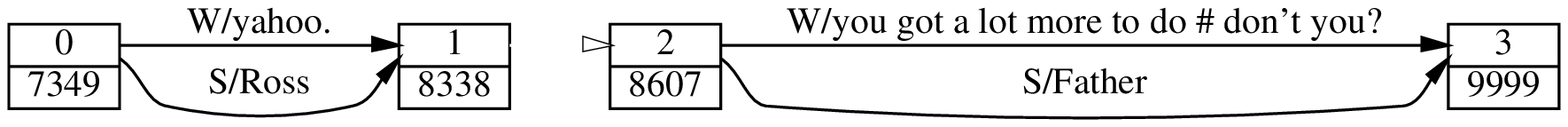,width=0.7\linewidth}}
\caption{Graph Structure for CHILDES Example (Version 1)}\label{chat2}
\vspace*{20ex}
\centerline{\epsfig{figure=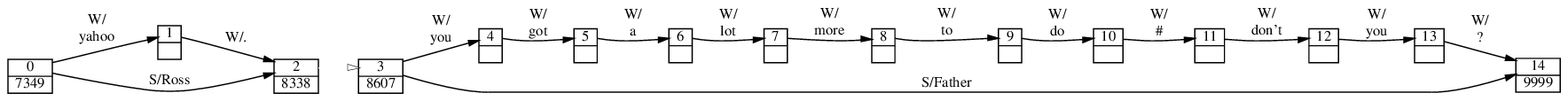,width=\linewidth}}
\caption{Graph Structure for CHILDES Example (Version 2)}\label{chat1}
\end{pagefig}

With its extensive user base, tools and documentation,
and its coverage of some two dozen languages,
the Child Language Data Exchange System, or CHILDES,
represents the largest scientific -- as opposed to
engineering -- enterprise involved in our survey.
The CHILDES database includes a vast amount of transcript data
collected from children and adults who are learning languages
\cite{MacWhinney95}.  
All of the data are transcribed in the so-called `CHAT' format;
a typical instance is provided by
this opening fragment of a CHAT transcription:

\begin{sv}
@Begin
@Filename:      boys73.cha
@Participants:  ROS Ross Child, MAR Mark Child,
                FAT Brian Father, MOT Mary Mother
@Date:  4-APR-1984
@Age of ROS:    6;3.11
@Sex of ROS:    Male
@Birth of ROS:  25-DEC-1977
@Age of MAR:    4;4.15
@Birth of MAR:  19-NOV-1979
@Sex of MAR:    male
@Situation:     Room cleaning
*ROS:   yahoo.
%snd:   "boys73a.aiff" 7349 8338
*FAT:   you got a lot more to do # don't you?
%snd:   "boys73a.aiff" 8607 9999
*MAR:   yeah.
%snd:   "boys73a.aiff" 10482 10839
*MAR:   because I'm not ready to go to
        <the bathroom> [>] +/.
%snd:   "boys73a.aiff" 11621 13784
\end{sv}

The \smtt{\%snd} lines, by the conventions of this notation, provide
times for the previous transcription lines, in milliseconds relative
to the beginning of the referenced file. The first two lines of this
transcript might then be represented graphically as in
Figure~\ref{chat2}.  Observe that the gap between the conversational
turns results in a disconnected graph.  Note also that the
\smtt{\%snd} annotations in the original chat file included a file
name; see \S\ref{sec:associations} for a discussion of associations
between annotations and files.

The representation in Figure~\ref{chat2} is inadequate, for it treats
entire phrases as atomic arc labels, complicating indexing and search.
We favor the representation in Figure~\ref{chat1}, where labels have
uniform ontological status regardless of the presence vs.\ absence of
time references.  Observe that most of the nodes in Figure~\ref{chat1}
{\it could} have been given time references in the CHAT format but
were not.  Our approach maintains the same topology regardless of the
sparseness of temporal information.

Notice that
some of the tokens of the transcript, i.e.\ the punctuation marks,
are conceptually not references to discrete stretches of time in the
same way that orthographic words are.  (The distinction could
be reflected by choosing a different type for punctuation labels.)
Evidently
it is not always meaningful to assign time references to the
nodes of an annotation.  We shall see a more pervasive example of this
atemporality in the next section.

\subsection{LACITO Linguistic Data Archiving Project}
\label{sec:lacito}

LACITO -- Langues et Civilisations \`a Tradition Orale -- is a CNRS
organization concerned with research on unwritten languages.  The
LACITO Linguistic Data Archiving Project was founded to conserve and
distribute the large quantity of recorded, transcribed speech data
collected by LACITO members over the last three decades
\cite{Michailovsky98}.  In this section we discuss a transcription for
an utterance in Hayu, a Tibeto-Burman language of Nepal.  The gloss
and free translation are in French.

\begin{sv}
<?XML version="1.0" encoding="ISO-8859-1" ?>
<!DOCTYPE ARCHIVE SYSTEM "Archive.dtd">

<ARCHIVE>
<HEADER>
  <TITLE>Deux soeurs</TITLE>
  <SOUNDFILE href="hayu.wav"/>
</HEADER>
<TEXT lang="hayu">
  <S id="s1">
    <TRANSCR>
      <W>nakpu</W>
      <W>nonotso</W>
      <W>si&#x014b;</W>
      <W>pa</W>
      <W>la&#x0294;natshem</W>
      <W>are.</W>
    </TRANSCR>
    <AUDIO type="wav" start="0.0000" end="5.5467"/>
    <TRADUC>On raconte que deux soeurs all&egrave;rent un jour chercher du bois.</TRADUC>
    <MOTAMOT>
      <W>deux</W>
      <W>soeurs</W>
      <W>bois</W>
      <W>faire</W>
      <W>all&egrave;rent(D)</W>
      <W>dit.on.</W>
    </MOTAMOT>
  </S>
</TEXT>
</ARCHIVE>
\end{sv}

\begin{pagefig}{15ex}
\centerline{\epsfig{figure=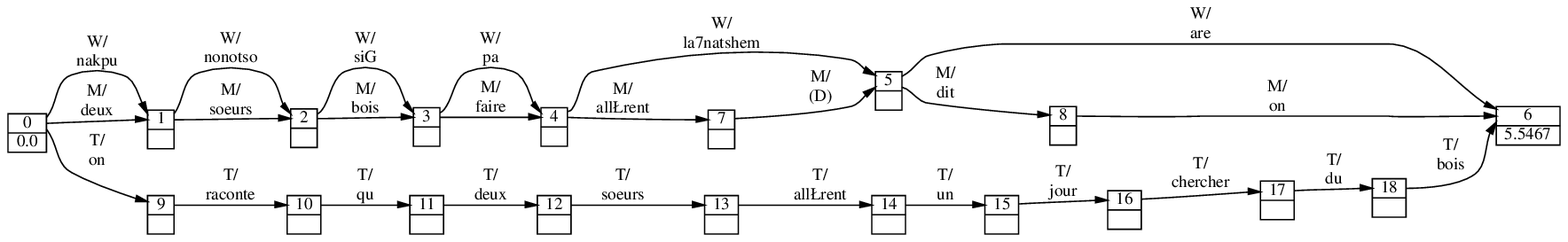,width=\linewidth}}
\caption{Graph Structure for the LACITO Archiving Project Example}\label{archivage}
\vspace*{15ex}
\centerline{\epsfig{figure=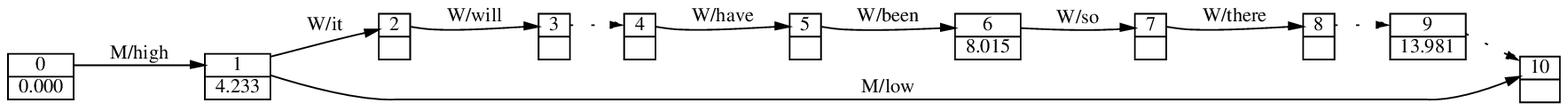,width=\linewidth}}
\caption{Graph Structure for LDC Broadcast Transcript Example}\label{ldc1}
\end{pagefig}

A possible graphical representation of the annotation of the sentence,
expressed as a labeled directed acyclic graph of the type under discussion, is
shown in Figure~\ref{archivage}.
Here we have three types of edge labels: \smtt{W/} for the words of the
Hayu story; \smtt{M/} for a word-by-word interlinear translation into
French; and \smtt{T/} for a phrasal translation into French.

(We have taken a small liberty with the word-by-word annotation in the
original file, which is arranged so that the \smtt{$<$W$>$} (for `word')
tokens in the Hayu are in one-to-one correspondence with the \smtt{$<$W$>$}
tokens in the French \smtt{$<$MOTAMOT$>$} interlinear version.  In such cases,
it is normal for individual morphemes in the source language to
correspond to several morphemes in the target language. This happens
twice in the sentence in question, and we have split the interlinear
translations to reflect the natural tokenization of the target
language.)

In this example, the time references (which are in seconds)
are again given only
at the beginning and end of the phrase, as required by the
LACITO Archiving Project format.  Nevertheless, the individual Hayu words have
temporal extent and one might want to indicate that in the annotation.
Observe that there is no meaningful way of assigning time
references to word boundaries in the phrasal translation.
Whether the time references happen to be unknown,
as in the upper half of Figure~\ref{archivage},
or are intrinsically un-knowable, as in the lower half of
Figure~\ref{archivage}, we can
treat the \smtt{W}, \smtt{M} and \smtt{T} annotations
in identical fashion.

\subsection{LDC Broadcast News Transcripts}

The Linguistic Data Consortium (LDC) is an open consortium of
universities, companies and government research laboratories, hosted
by the University of Pennsylvania, that creates, collects and
publishes speech and text databases, lexicons, and similar resources.
Since its foundation in 1992, it has published some 150 digital
databases, most of which contain material that falls under our
definition of `linguistic annotation.'

The Hub-4 English broadcast news corpora from the LDC
contain some 200 hours of speech data with SGML annotation
[\smtt{www.ldc.upenn.edu/Catalog/LDC\{97T22,98T28\}.html}].
About 60 hours of similar material has been published in
Mandarin and Spanish, and an additional corpus of some 700 hours of
English broadcast material will be published this year.
What follows is the beginning of a radio program transcription from
the Hub-4 corpus.

\begin{sv}
<Background Type=Music Time=0.000 Level=High>
<Background Type=Music Time=4.233 Level=Low>
<Section S_time=4.233 E_time=59.989 Type=Filler>

<Segment S_time=4.233 E_time=13.981 Speaker="Tad_Bile" Fidelity=Low Mode=Spontaneous>
    it will certainly make some of these districts more competitive than they have been
  <Sync Time=8.015>
    so there will be some districts which are republican
  <Sync Time=11.040>
    but all of a sudden they may be up for grabs
</Segment>
<Segment S_time=13.981 E_time=40.840 Speaker="Noah_Adams" Fidelity=High Mode=Planned>
    politicians get the maps out again
  <Sync Time=15.882>
    for friday june fourteenth this is n. p. r.'s all things considered
  <Sync Time=18.960>
  <Background Type=Music Time=23.613 Level=Low>
  <Sync Time=23.613>
    in north carolina and other states officials are trying to figure out the
    effects of the supreme court ruling against minority voting districts {breath}
  <Sync Time=29.454>
    a business week magazine report of a federal criminal investigation {breath}
  <Sync Time=33.067>
    into the cause and the aftermath of the ValuJet crash in florida {breath}
  <Sync Time=36.825>
    efforts in education reform {breath} and the question will the public pay
</Segment>
\end{sv}

Transcriptions are divided into sections (see the \smtt{Section} tag),
where each section consists
of a number of \smtt{Segment} blocks.
At various times during a segment
a \smtt{Sync Time} element is inserted to align a word boundary
with an offset into a speech file.
Elements specifying changes in background noise and signal quality
function independently of the hierarchy.  For example, a period
of background music might bridge two segments, beginning
in one segment and ending in the next.  Figure~\ref{ldc1}
represents the structure of this annotation.  Dotted arcs
represent elided material, \smtt{W/} is for words and
\smtt{M/} is for background music level.

\subsection{LDC Telephone Speech Transcripts}
\label{sec:callhome}

The LDC-published CALLHOME corpora include digital audio, 
transcripts and lexicons for telephone conversations in several languages.
The corpora are designed to support research on speech recognition algorithms
[\smtt{www.ldc.upenn.edu/Catalog/LDC96S46.html}].
The transcripts exhibit abundant overlap between speaker turns
in two-way telephone conversations.

What follows is a typical fragment of an annotation.  Each stretch of
speech consists of a begin time, an end time, a speaker designation
(`A' or `B' in the example below), and the transcription for the
cited stretch of time.  We have augmented the annotation with \smtt{+}
and \smtt{*} to indicate partial and total overlap (respectively) with
the previous speaker turn.

\begin{sv}
  962.68 970.21 A: He was changing projects every couple of weeks and he
  said he couldn't keep on top of it. He couldn't learn the whole new area  
* 968.71 969.00 B: %mm.  
  970.35 971.94 A: that fast each time.  
* 971.23 971.42 B: %mm.    
  972.46 979.47 A: %um, and he says he went in and had some tests, and he
  was diagnosed as having attention deficit disorder. Which  
  980.18 989.56 A: you know, given how he's how far he's gotten, you know,
  he got his degree at &Tufts and all, I found that surprising that for
  the first time as an adult they're diagnosing this. %um  
+ 989.42 991.86 B: %mm. I wonder about it. But anyway.  
+ 991.75 994.65 A: yeah, but that's what he said. And %um  
* 994.19 994.46 B: yeah.  
  995.21 996.59 A: He %um  
+ 996.51 997.61 B: Whatever's helpful.  
+ 997.40 1002.55 A: Right. So he found this new job as a financial
  consultant and seems to be happy with that.  
  1003.14 1003.45 B: Good.  
+ 1003.06 1006.27 A: And then we saw &Leo and &Julie at Christmas time.  
* 1005.45 1006.00 B: uh-huh.    
  1006.70 1009.85 A: And they're doing great. %um, they had just moved to  
+ 1009.25 1010.58 B: He's in &New &York now, right?  
+ 1010.19 1013.55 A: a really nice house in &Westchester. yeah, an o-  
+ 1013.38 1013.61 B: Good.  
+ 1013.52 1018.57 A: an older home that you know &Julie is of course
  carving up and making beautiful. %um  
* 1018.15 1018.40 B: uh-huh.  
  1018.68 1029.75 A: Now she had a job with an architectural group
  when she first got out to &New &York, and that didn't work out. She
  said they had her doing things that she really wasn't qualified to do  
\end{sv}

Long turns (e.g.\ the period from 972.46 to 989.56 seconds) were
broken up into shorter stretches for the convenience of the
annotators.  Thus this format is ambiguous as to whether adjacent
stretches by the same speaker should be considered parts of the same
unit, or parts of different units -- in translating to an annotation
graph representation, either choice could be made. However, the intent
is clearly just to provide additional time references within long
turns, so the most appropriate choice seems to be to merge abutting
same-speaker structures while retaining the additional time-marks.

A section of this annotation including an example of total
overlap is represented in annotation graph form in Figure~\ref{callhome}.
The turns are attributed to speakers using the \smtt{speaker/} type.
All of the words, punctuation and disfluencies are given the \smtt{W/}
type, though we could easily opt for a more refined version in which
these are assigned different types.
Observe that the annotation graph representation preserves the
non-explicitness of the original file format concerning which
of speaker A's words overlap which of speaker B's words. Of course,
additional time references could specify the overlap down to any
desired level of detail (including to the level of phonetic segments
or acoustic events if desired).

\subsection{NIST Universal Transcription Format}

The US National Institute of Standards and Technology (NIST) has
recently developed a set of annotation conventions `intended to
provide an extensible universal format for transcription and
annotation across many spoken language technology evaluation domains'
\cite{UTF98}.  This `Universal Transcription Format' (UTF) was based
on the LDC Broadcast News format, previously discussed. A key design
goal for UTF was to provide an SGML-based format that would cover both
the LDC broadcast transcriptions and also various LDC-published
conversational transcriptions, while also providing for plausible
extensions to other sorts of material.

A notable
aspect of UTF is its treatment of overlapping speaker turns.
In the following fragment (from the Hub-4 1997 evaluation set),
overlapping stretches of speech are marked with the
\smtt{$<$b\_overlap$>$} (begin overlap) and
\smtt{$<$e\_overlap$>$} (end overlap) tags.

\begin{sv}
<turn speaker="Roger_Hedgecock" spkrtype="male" dialect="native"
    startTime="2348.811875" endTime="2391.606000" mode="spontaneous" fidelity="high">
  ...
  <time sec="2378.629937">
  now all of those things are in doubt after forty years of democratic rule in
  <b_enamex type="ORGANIZATION">congress<e_enamex>
  <time sec="2382.539437">
  \{breath because <contraction e_form="[you=>you]['ve=>have]">you've got quotas
  \{breath and set<hyphen>asides and rigidities in this system that keep you
  <time sec="2387.353875">
  on welfare and away from real ownership
  \{breath and <contraction e_form="[that=>that]['s=>is]">that's a real problem in this
  <b_overlap startTime="2391.115375" endTime="2391.606000">
    country
  <e_overlap>
</turn>
<turn speaker="Gloria_Allred" spkrtype="female" dialect="native"
    startTime="2391.299625" endTime="2439.820312" mode="spontaneous" fidelity="high">
  <b_overlap startTime="2391.299625" endTime="2391.606000">
    well i
  <e_overlap>
  think the real problem is that %uh these kinds of republican attacks
  <time sec="2395.462500">
  i see as code words for discrimination
  ...
</turn>
\end{sv}

\begin{pagefig}{0ex}
\centerline{\epsfig{figure=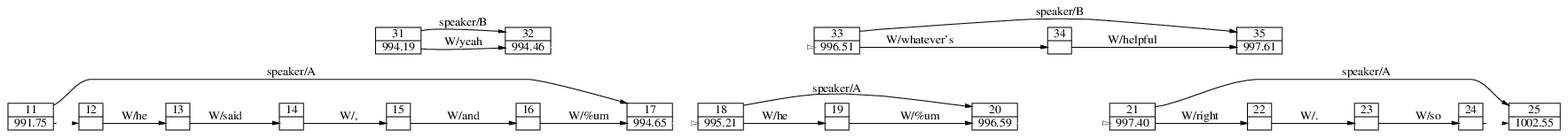,width=\linewidth}}
\caption{Graph Structure for LDC Telephone Speech Example}\label{callhome}
\vspace*{8ex}
\centerline{\epsfig{figure=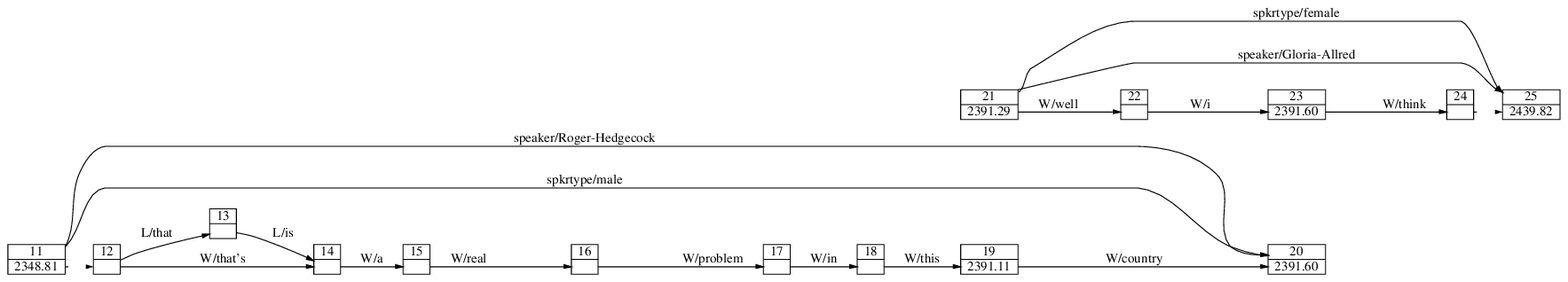,width=\linewidth}}
\caption{Graph Structure for NIST UTF Example}\label{utf}
\vspace*{8ex}
\centerline{\epsfig{figure=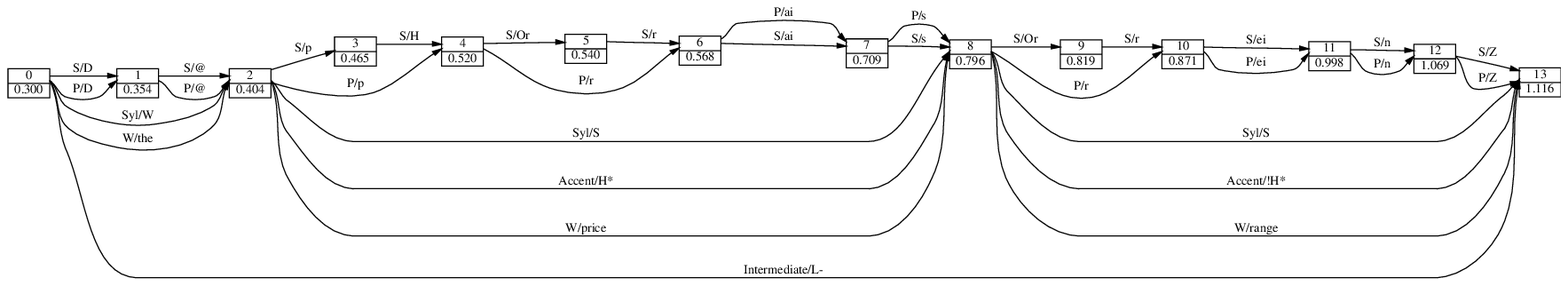,width=\linewidth}}
\caption{Graph Structure for Emu Annotation Example}\label{emu1}
\end{pagefig}

Observe that there are two speaker turns, where the first speaker's
utterance of `country' overlaps the second speaker's utterance of
`well I'.  Note that the time attributes for overlap are not required
to coincide, since they are aligned to `the most inclusive word
boundaries for each speaker turn involved in the overlap'.  The
coincidence of end times in this case is almost surely an artifact of the user
interface of the system used to create the annotations, which
required overlaps to be specified relative to word boundaries.

The structure of overlapping turns can be represented using annotation
graphs as shown in Figure~\ref{utf}.  Each speaker turn is a separate
connected subgraph, disconnected from other speaker turns.
This situation neatly reflects the fact that
the time courses of utterances by various speakers in
conversation are logically asynchronous.
Observe that the information
about overlap is implicit in the time references and
that partial word overlap can be represented. This seems like the
best choice in general, since there is no necessary logical structure
to conversational overlaps -- at base, they are just two different
actions unfolding over the same time period.

The cited annotation graph structure is thus less explicit about word
overlaps than the UTF file.  However, if a more explicit symbolic
representation of overlaps is desired, specifying that such-and-such a
stretch of one speaker turn is associated with such-and-such a stretch
of another speaker turn, this can be represented in our framework
using the inter-arc linkage method described in \S\ref{sec:multiple},
or using the extension described in \S\ref{sec:extensions}.

Of course, the same word-boundary-based representation of overlapping
turns could also be expressed in annotation graph form, by allowing
different speakers' transcripts to share certain nodes (representing
the word boundaries at which overlaps start or end). We do not suggest this,
since it seems to us to be based on an inappropriate model of overlapping,
which will surely cause trouble in the end.

Note the use of the \smtt{L/} `lexical' type to include the
full form of a contraction.  The UTF format employed special
syntax for expanding contractions.  No additional ontology
was needed in order to do this in the annotation graph.
(A query to find instances of \smtt{W/that} or \smtt{L/that}
would simply disjoin over the types.)

Note also that it would have been possible to replicate the type
system, replacing \smtt{W/} with \smtt{W1/} for `speaker 1'
and \smtt{W2/} for `speaker 2'.  However, we have chosen instead
to attribute material to speakers using the \smtt{speaker/}
type on an arc spanning an entire turn.  The disconnectedness
of the graph structure means there can be no ambiguity about
the attribution of each component arc to a speaker.

As we have argued, annotation graphs of the kind shown in Figure~\ref{utf} are
actually more general and flexible than the UTF files they
model.  The UTF format imposes a linear structure on the speaker
turns and assumes that overlap only occurs at the periphery of
a turn.  In contrast, the annotation graph structure is well-behaved
for partial word overlap, and it scales
up naturally and gracefully to the situation where multiple speakers are
talking simultaneously (e.g.\ for transcribing a radio talk-back
show with a compere, a telephone interlocutor and a panel of discussants).
It also works for arbitrary kinds of overlap (e.g.\ where one speaker turn
is fully contained inside another), as discussed in the previous section.

\subsection{Emu}

The Emu speech database system \cite{Cassidy96}
grew out of the earlier Mu+ (Macquarie University)
system \cite{Harrington93}, which was designed to
support speech scientists who work with large collections
of speech data, such as the
Australian National Database of Spoken Language
[\smtt{andosl.anu.edu.au/andosl}].

Emu permits hierarchical annotations arrayed over any
number of levels, where each level is a linear ordering.
An annotation resides in a single file linked
to an xwaves label file.  The file begins with a declaration
of the levels of the hierarchy and the
immediate dominance relations.

\begin{sv}
level Utterance
level Intonational      Utterance
level Intermediate      Intonational
level Word              Intermediate
level Syllable          Word
level Phoneme           Syllable
level Phonetic          Phoneme    many-to-many
\end{sv}

The final line licenses a many-to-many relationship
between phonetic segments and phonemes, rather than
the usual many-to-one relationship.
According to the user's manual,
this is only advisable at the bottom of the hierarchy,
otherwise temporal ambiguities may arise.

At any given level of the hierarchy, the elements may have more than one
attribute.  For example, in the following declarations we see
that elements at the \smtt{Word} level may be decorated with
\smtt{Accent} and \smtt{Text} information, while syllables may
carry a pitch accent.

\begin{sv}
label Word              Accent
label Word              Text
label Syllable          Pitch_Accent
\end{sv}

The next line sets up a dependency between the \smtt{Phonetic} level
and an xwaves label file linked to ESPS-formatted audio data.

\begin{sv}
labfile Phonetic :format ESPS :type SEGMENT :mark END :extension lab :time-factor 1000
\end{sv}

The \smtt{type} declaration distinguishes
`segments' with duration from `events' which
are instantaneous.
Here, the time associated with
a segment will mark its endpoint rather than its
starting point, as indicated by the \smtt{mark END} declaration.
The timing information from the label file is adopted
into the hierarchy (scaled from $\mu$s to ms),
and can propagate upwards.  In this way, the end of
a phonetic segment may also become the end of a syllable,
for example.

The sequence of labels from the xwaves label file is reproduced in the
Emu annotation, while the timing information remains in
the xwaves label file.  Therefore the latter file is an essential
part of an Emu annotation and must be explicitly referenced.
The labels are assigned unique numerical identifiers, as shown below
for the sentence `the price range is smaller than any of us
expected'.  (For compactness, multiple lines have been
collapsed to a single line.)

\begin{sv}
Phonetic Phonetic
0 D     9 @     11 p    16 H    17 Or    19 r    20 ai   22 s    24 Or
30 r    31 ei   33 n    35 Z    37 I     44 zs   50 Om   52 m    53 o:
55 l    58 @    60 D    65 @    67 n     69 EC   76 E    77 n    80 i:
82 @    88 v    90 @    95 s    97 I     102 k   104 H   105 s   109 p
111 H   112 E   114 k   116 H   117 t    120 H   121 @   123 d   125 H
\end{sv}

The labels on the more abstract, phonemic level are assigned a
different set of numerical identifiers.

\begin{sv}
Phoneme Phoneme
1 D     10 @    12 p    18 r    21 ai   23 s    25 r    32 ei   34 n
36 Z    38 I    45 z    46 s    51 m    54 o:   56 l    59 @    61 D
66 @    68 n    70 E    78 n    81 i:   83 @    89 v    91 @    96 s
98 I    103 k   106 s   110 p   113 E   115 k   118 t   122 @   124 d
\end{sv}

Here is the remainder of the hierarchy.
 
\begin{sv}
Utterance Utterance
8

Intonational Intonational
7 L%

Intermediate Intermediate
5 L-      42 L-      74 L-

Word Word Accent Text
2 F W the             13 C S price         26 C S range          39 F W is
47 C S smaller        62 F W than          71 F S any            84 F W of
92 F W us             99 C S expected

Syllable Syllable Pitch_Accent
4 W        15 S H*    28 S !H*    41 W      49 S H*    57 W       64 W
73 S       79 W H*    86 W        94 W      101 W      108 S H*   119 W
\end{sv}

A separate section of an Emu annotation file
lists each identifier, followed by all those
identifiers which it dominates.
For example,
the line \smtt{4 0 1 9 10} states that the first \smtt{W}
syllable (id=4) directly or indirectly dominates
phonetic segments \smtt{D} (id=0) and \smtt{@} (id=9)
and phonemes \smtt{D} (id=1) and \smtt{@} (id=10).
The first intermediate phrase label \smtt{L-} (id=5)
dominates this material and much other material besides:

\begin{sv}
5 0 1 2 4 9 10 11 12 13 15 16 17 18 19 20 21 22 23 24 25 26 28 30 31 32 33 34 35 36
\end{sv}

This exhaustive approach greatly facilitates the display of parts of
the annotation hierarchy.  If the syllable level is switched
off, it is a trivial matter to draw lines directly from words to phonemes.

The first three words of this annotation are displayed as an
annotation graph in Figure~\ref{emu1}.  Here \smtt{S/} is
used for phonetic segments, \smtt{P/} for phonemes and
\smtt{Syl/} for strong (\smtt{S}) and weak (\smtt{W}) syllables.

\subsection{Festival}\label{sec:festival}

The Festival speech synthesis system \cite{Taylor98,Taylor99}
is driven by richly-structured linguistic input.
The Festival data structure, called a `heterogeneous relation
graph' (HRG) is a collection of binary relations
over attribute-value matrices (AVMs).  Each matrix
describes the local properties of some linguistic unit,
such as a segment, a syllable, or a syntactic phrase.
The value of an attribute could be atomic (such as a binary
feature or a real number), or another (nested) AVM, or a function.
Functions have the ability to traverse one or more binary relations
and incorporate values from other AVMs.
For example, if duration was an attribute of a syllable, its
value would be a function subtracting the start time of the
first dominated segment from the end time of the last dominated
segment.  Typically, each level of structure includes these
function-valued attributes so that temporal information
is correctly propagated and does not need to be stored more
than once.

\begin{pagefig}{0ex}
\centerline{\epsfig{figure=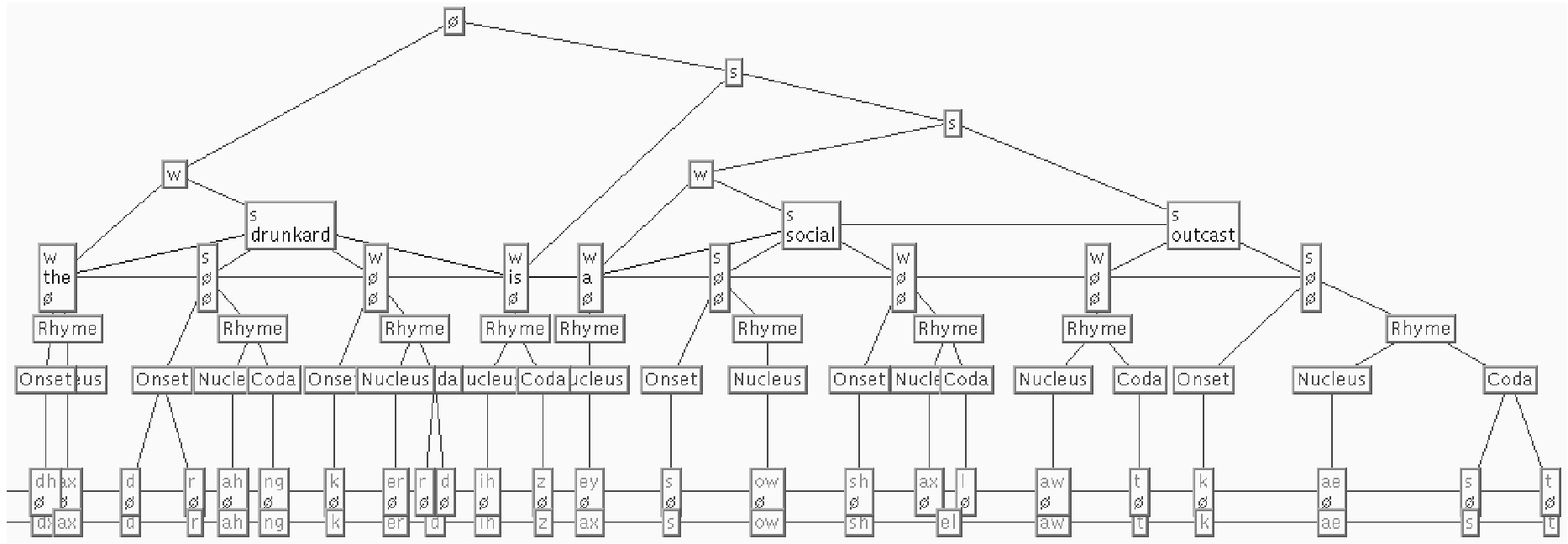,width=0.9\textheight}}
\caption{Annotation Structure from Festival}\label{taylor}
\vspace*{5ex}
\centerline{\epsfig{figure=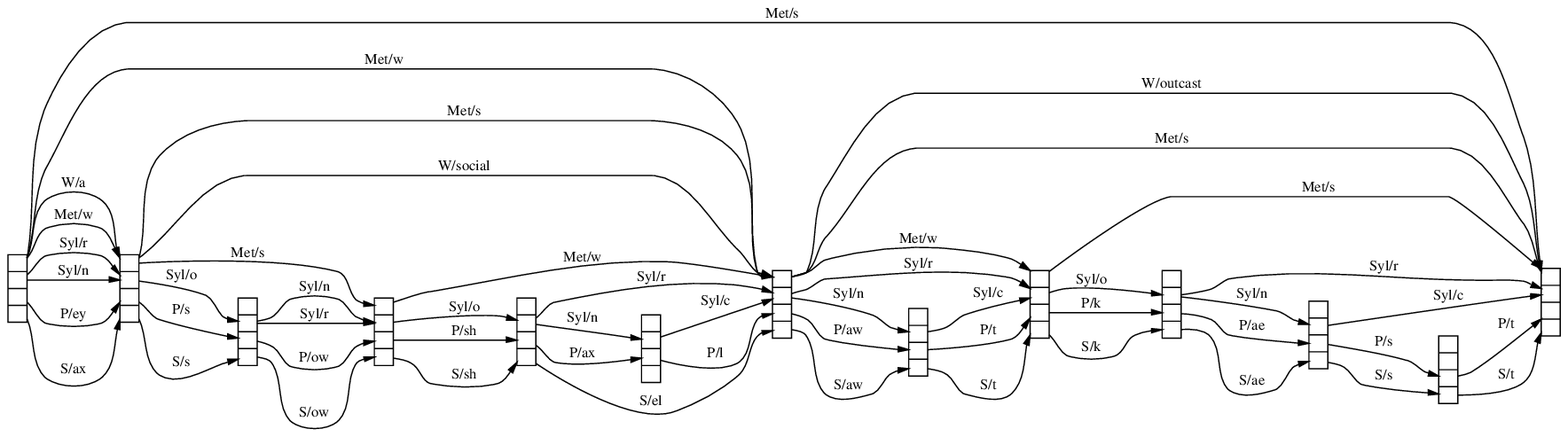,width=0.9\textheight}}
\caption{Graph Structure for Festival Example}\label{festival}
\end{pagefig}

An example HRG is shown in Figure~\ref{taylor}.
Each box contains an abbreviated form of an AVM.  The lines
represent the binary relations.  Observe, for example, that the
phonemes and the surface segments are organized into two sequences,
the two parallel lines spanning the bottom of the figure.  Each
sequence is a distinct binary relation.  The hierarchical structures
of the metrical and the syllable trees are two more binary relations.
And the linear ordering of words is still another binary relation.

Figure~\ref{festival} gives the annotation graph representing the
second half of the HRG structure.  Given the
abundance of arcs and levels, we have expanded the vertical
dimension of the
nodes, but this is not significant.  Node identifiers and time
references have been omitted.
Like the HRG, the annotation graph represents temporal
information only once.  Yet unlike the HRG, there is no
need to define explicit propagation functions.

\section{Architectural Considerations}
\label{sec:arch}

A diverse range of annotation models have now been considered.
Our provision of annotation graphs for each one already gives a
foretaste of the formalism we present in \S\ref{sec:algebra}.
However, before launching into the formalism, we want to stand
back from the details of the various models, and try to take in the
big picture.  In this section we describe a wide
variety of architectural issues which we believe should be addressed
by any general purpose model for annotating linguistic signals.

\subsection{Representation of Partial Information}

In the discussion of CHILDES and the LACITO Archiving Project
above, there were cases where our graph representation
had nodes which bore no time reference.
Perhaps times were not measured, as in typical annotations
of extended recordings where time references might
only be given at major phrase boundaries (c.f.\ CHILDES).
Or perhaps time measurements were
not applicable in principle, as for phrasal translations
(c.f.\ the LACITO Archiving Project).
Various other possibilities suggest themselves.
We might create a segment-level annotation automatically from a word-level
annotation by looking up each word in a pronouncing dictionary
and adding an arc for each segment, prior to hand-checking
the segment annotations and adding time references to the
newly created nodes.  The annotation should remain well-formed
(and therefore usable) at each step in this enrichment process.

Just as the temporal information may be partial, so might
the label information.  For example, we might label indistinct
speech with whatever information is available -- `so-and-so said something
here that seems to be two syllables long and begins with a /t/'.

Beyond these two kinds of partiality, there is an even more
obvious kind of partiality we should recognize.
An annotated corpus might be annotated in a fragmentary
manner.  It might be that only 1\% of a certain recording has any bearing
on the research question that motivated the collection and annotation
work.  Therefore, it should be possible
to have a well-formed annotation structure
with arbitrary amounts of annotation detail at certain
interesting loci, and limited or no detail elsewhere.
This is a typical situation in phonetic or sociolinguistic research,
where a large body of recordings may be annotated in detail with
respect to a single, relatively infrequent phenomenon of interest.

Naturally, one could always extract a sub-corpus and annotate
that material completely, thereby removing the need for partiality,
but this may have undesirable consequences for managing a corpus:
(i) special intervention is required each time one wants
to expand the sub-corpus as the research progresses;
(ii) it is difficult to make annotations of a sub-corpus
available to someone working on a related research question
with an overlapping sub-corpus, and updates cannot be
propagated easily;
(iii) provenance issues arise, e.g.\ it may be difficult to
identify the origin of any given fragment, in case access to
broader context is necessary to retrieve the value of some
other independent variable one might need to know; and
(iv) it is difficult to combine the various contributions
into the larger task of annotating a standard corpus for
use in perpetuity.

By pointing out these problems we do not mean to suggest that
all annotations of a corpus should be physically or logically combined.
On the contrary, even with one physical copy of a corpus, we would want
to allow several independent (partial) annotations to coexist,
where these may be owned by different
people and stored remotely from each other.  Nor do we wish to
suggest that the creation of sub-corpora is never warranted.  The
point is simply that an annotation formalism should not force users
to create a derived corpus just so that a partial annotation is
well-formed.

\subsection{Encoding Hierarchical Information}\label{sec:hierarchy}

Existing annotated speech corpora always involve a hierarchy of
several levels of annotation, even if they do not focus on very
elaborate types of linguistic structure.  TIMIT has sentences, words
and phonetic segments; a broadcast news corpus may have designated
levels for shows, stories, speaker turns, sentences and words. 

Some annotations may express much more elaborate hierarchies, with
multiple hierarchies sometimes created for a single underlying body of
speech data. For example, the Switchboard corpus of conversational
speech \cite{Godfrey92} began with the three basic levels: conversation,
speaker turn, and word. Various parts of it have since been annotated
for syntactic structure \cite{Marcus93}, for breath groups and
disfluencies \cite{Taylor95}, for speech act type
\cite{JuretafskyBates97,JurafskyShriberg97}, and for phonetic
segments \cite{Greenberg96}. These various annotations have been done as separate
efforts, and presented in formats that are fairly easy to process one-by-one,
but difficult to compare or combine.

Considering the variety of approaches that have been adopted,
it is possible to identify at least three general methods
for encoding hierarchical information.

\begin{description}
\item[Token-based hierarchy]
Here, hierarchical relations among annotations are explicitly
marked with respect to particular tokens: `this
particular segment is a daughter of this particular syllable.'
Systems that have adopted this approach include
Partitur, Emu and Festival. 

\item[Type-based hierarchy] 
Here, hierarchical information is given with respect to types --
whether once and for all in the database, or ad hoc by a user, or
both. In effect, this means that a grammar of some sort is specified,
which induces (additional) structure in the annotation.
This allows (for instance) the subordination of syllables to
words to be indicated, but only as a general fact about all syllables
and words, not as a specific fact about particular syllables and
words.  An SGML DTD is an example of this: it specifies a context-free
grammar for any textual markup that uses it. In some cases, the
hierarchical structure of a particular stretch of SGML markup cannot
be determined without reference to the applicable DTD.

\item[Graph-based hierarchy]

Here, annotations are akin to the arcs in so-called `parse charts'
\cite[179ff]{GazdarMellish89}. A parse chart is a particular kind of
acyclic digraph, which starts with a string of words and then adds a
set of arcs representing hypotheses about constituents dominating
various substrings. In such a graph, if the substring spanned by arc
$a_i$ properly contains the substring spanned by arc $a_j$, then the
constituent corresponding to $a_i$ must dominate the constituent
corresponding to $a_j$ (though of course other structures may
intervene).  Hierarchical relationships are encoded in a parse chart
only to the extent that they are implied by this graph-wise inclusion
-- thus two arcs spanning the same substring are unspecified as to
their hierarchical relationship, and arcs ordered by temporal
inclusion acquire a hierarchical relationship even when this
is not appropriate given the types of those arcs
(though a grammar, external to the
parse chart for a particular sentence, may settle the matter;
see also \S\ref{sec:hierarchy-local}).

As we have seen, many sorts of linguistic annotations are naturally
encoded as graph structures with labeled arcs and time-marked
nodes. Such a representation arises naturally from the fact that
elementary annotations are predicates about stretches of
signal. Thus in our TIMIT example, we can construe the underlying
sequence of audio samples as a sort of terminal string, with
annotations representing hypotheses about constituents of various
types that dominate designated subsequences. In the example cited, the
word `she' spans the sequence from sample 2360 to sample 5200; the
phoneme /sh/ spans the sequence from 2360 to 3720; and the phoneme
/iy/ spans the sequence from 3720 to 5200. This graph structure itself
implies a sort of hierarchical structure based on temporal inclusion.
If we interpret it as a
parse chart, it tells us that the word `she' dominates the phoneme
sequence /sh~iy/.  Examples of annotation systems that encode
hierarchy using this approach are TIMIT, CHILDES and Delta
\cite{Hertz90}.  (Note that, once equipped with the full annotation
graph formalism, we will be able to distinguish graph-based and
time-based inclusion, conflated here.)

\end{description}

A particular system may present some mixture of the above techniques.
Thus an SGML labeled bracketing may specify an unambiguous
token-based hierarchy, with the applicable DTD grammar being just a
redundant type-based check; but in some cases, the DTD may be
necessary to determine the structure of a particular stretch of
markup. Similarly, the graph structures implicit in TIMIT's annotation
files do not tell us, for the word spelled `I' and pronounced /ay/,
whether the word dominates the phoneme or vice versa; but the
structural relationship is implicit in the general relationship
between the two types of annotations.

An annotation framework (or its implementation) may also choose to
incorporate arbitrary amounts of redundant encoding of structural
information. It is often convenient to add redundant links explicitly
-- from children to parents, from parents to children, from one child
to the next in order, and so on -- so that a program can navigate the
structure in a way that is clearer or more efficient.  Although such
redundant links can be specified in the basic annotation itself -- as
in {\em Festival} -- they might equally well be added
automatically, as part of a compilation or indexing process.  In our
view, the addition of this often-useful but predictable structure
should not be an intrinsic part of the definition of general-purpose
annotation structures.  We want to distinguish the annotation
formalism itself from various enriched data structures with redundant
encoding of hierarchical structure, just as we would distinguish it
from various indices for convenient searching of labels and label
sequences.

In considering how to encode hierarchical information, we start from
the premise that our representation will include some sort of graph
structure, simply because this is the most fundamental and natural sort of
linguistic annotation. Given this approach, hierarchical structure can
often be read off the annotation graph structure, as was suggested
informally above and will be discussed more thoroughly in
\S\ref{sec:algebra}. For many applications, this will be enough. For the
residual cases, we might add either type-based or token-based encoding
of hierarchical information (see \S\ref{sec:extensions}).

Based on the formal precedent of SGML, the model of how chart-like
data structures are actually used in parsing, and the practical
precedents of databases like TIMIT, it is tempting to consider adding
a sort of grammar over arc labels as part of the formal definition of
annotation graphs. However, in the absence of carefully-evaluated
experience with circumstances in which this move is motivated, we
prefer to leave this as something to be added by particular applications
rather than incorporated into the formalism.
In any case, we shall argue later (see \S\ref{sec:multiple})
that we need a more general method to encode optional relationships
among particular arcs. This method permits token-based marking of
hierarchical structure as a special case.

We also need to mention that particular applications in
the areas of creation, query and display of annotations may
be most naturally organized in ways that motivate a user interface
based on a different sort
of data structure than the one we are proposing.
For instance, it may sometimes be easier to create
annotations in terms of tree-like dominance relations rather than chart-like
constituent extents, for instance in doing syntactic tree-banking
\cite{Marcus93}. It may likewise be easier in some cases to define 
queries explicitly in terms of tree structures. And finally, it may
sometimes be more helpful to display trees rather than equivalent
annotation graphs -- the Festival example in \S\ref{sec:festival} was 
a case in point. We believe that such user interface issues will
vary from application to application, and may even depend on the
tastes of individuals in some cases. In any case, decisions about
such user interface issues are separable from decisions about
the appropriate choice of basic database structures.

\subsection{Gestural scores and multiple nodes at a time point}\label{sec:scores}

\begin{figure}
\centerline{\epsfig{figure=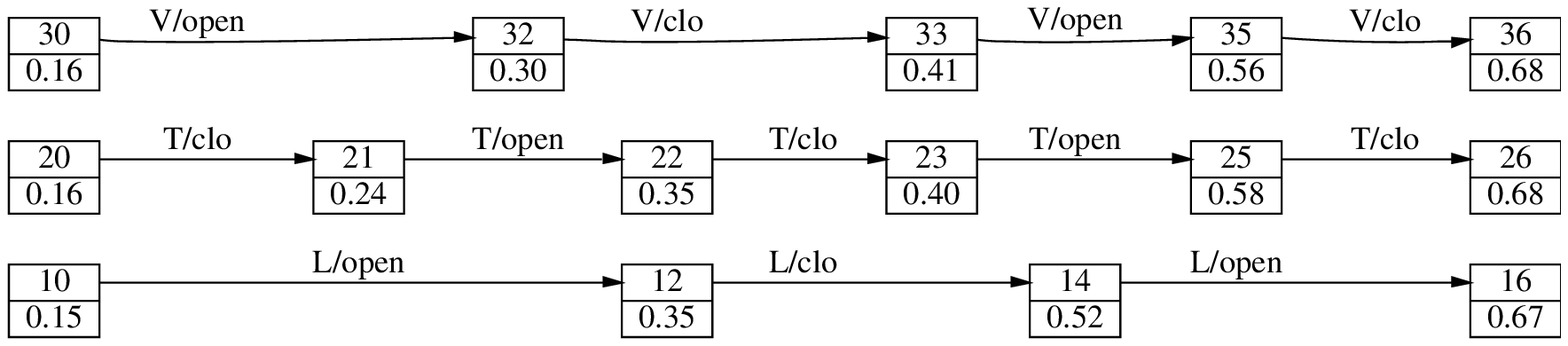,width=\linewidth}}
\caption{Gestural Score for the Phrase 'ten pin'}\label{tenpin3}
\vspace*{2ex}\hrule
\end{figure}

In addition to the hierarchical and sequential structuring
of information about linguistic signals, we also have parallel
structuring.  Nowhere is this clearer than in the gestural
score notation used to describe the articulatory component
of words and phrases (e.g.\ \cite{Browman89}).
A gestural score maps out the time course of the gestural
events created by the articulators of the vocal tract.
This representation expresses the fact that the articulators
move independently and that the segments we observe are the
result of particular timing relationships between the gestures.
Figure~\ref{tenpin3} gives the annotation graph for a gestural score.
It shows the activity
of the velum \smtt{V/}, the tongue tip \smtt{T/} and the
lips \smtt{L/}.
This example stands in stark contrast to the hierarchical
structures discussed in the previous section.  Here there
is no hierarchical relationship between the streams.

Another important difference between hierarchical and
parallel structures needs to be drawn here.
Suppose that two labeled periods of an annotation begin (or end)
at the same time.  The alignment of two such boundaries might
be necessary, or pure coincidence.
As an example of necessary alignment, consider the case of
phrase-initial words.  Here, the left boundary of a
phrase lines up with the left boundary of its initial word.
Changing the time of the phrase boundary should change the
time of the word boundary, and vice versa.
In the general case, an update of this sort must propagate
both upwards and downwards in the hierarchy.
In fact, we argue that these two pieces of annotation actually
{\em share} the same boundary: their arcs emanate from a single
node.  Changing the time reference of that node does not
need to propagate anywhere, since the information is already
shared by the relevant arcs.

As an example of coincidental alignment, consider the
case of gestural scores once more.  In 100 annotated recordings
of the same utterance we might find that the boundaries of
different gestures occasionally coincide.  An example of
this appears in Figure~\ref{tenpin3}, where nodes 12 and
22 have the same time reference.  However, this alignment
is a contingent fact about a particular utterance token.
An edit operation which changed the start time of one gesture would
usually carry no implication for the start time of some
other gesture.

\subsection{Instants, overlap and gaps}
\label{sec:instants}

Even though a linguistic event might have duration, such as the attainment
of a pitch target, the most perspicuous annotation may be tied
to an instant rather than an interval.  Some annotation
formalisms (e.g.\ Emu, Festival, Partitur)
provide a way to label instants.
The alignment of these instants with respect to other instants
or intervals can then be investigated or exploited.
There are at least five conceivable approaches to
labeled instants (note that this is not a mutually exclusive set):

\begin{enumerate}
\item
  nodes could be optionally labeled; or
\item
  an instant can be modeled as a self-loop on a node,
  and again labeled just like any other arc; or
\item
  instants can be treated as arcs between two nodes
  with the same time reference; or
\item
  instants can be treated as short periods, where
  these are labeled arcs just like any other; or
\item
  certain types of labels on periods could be interpreted
  as referring to the commencement or the culmination of that period.
\end{enumerate}

With little evidence on which to base a decision between
these options we opt for the most conservative, which is
the one embodied in the last two options.  Thus with no extension to
the ontology we already have two ways to model instants.

\begin{figure}[t]
\centerline{\epsfig{figure=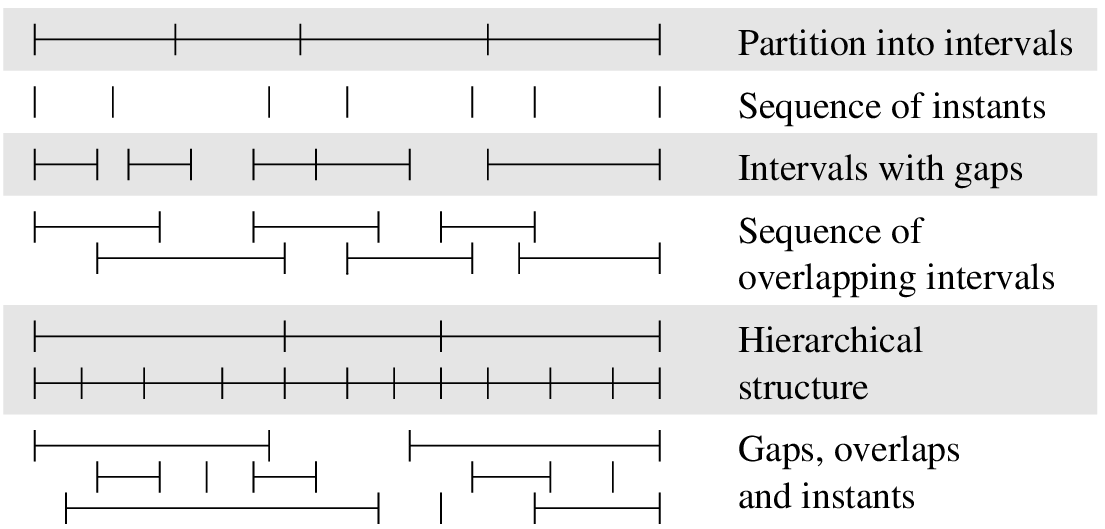,width=0.85\linewidth}}
\caption{Possible Structures for a Single Layer}\label{layer}
\vspace*{2ex}\hrule
\end{figure}

As we have seen, annotations are often stratified, where each
layer describes a different property of a signal.
What are the possible temporal relationships between
the pieces of a given layer?  Some possibilities are
diagrammed in Figure~\ref{layer}, where a point is
represented as a vertical bar, and an interval is represented
as a horizontal line between two points.

In the first row of Figure~\ref{layer}, we see a layer which
exhaustively partitions the time-flow into a sequence
of non-overlapping intervals (or perhaps intervals which
overlap just at their endpoints).  In the second row we see
a layer of discrete instants.  The next two rows illustrate
the notions of gaps and overlaps.  Gaps might correspond to
periods of silence, or to periods in between the salient events,
or to periods which have yet to be annotated.
Overlaps occur between speaker turns in discourse (see Figure~\ref{callhome})
or even between adjacent words in a single speech stream
(see Figure~\ref{links}a).
The fifth row illustrates a hierarchical grouping of intervals
within a layer (c.f. the \smtt{Met/} arcs in Figure~\ref{festival}).
The final row contains an arbitrary set of intervals and
instants.

We adopt this last option (minus the instants)
as the most general case for the layer of an annotation.
As we shall see, layers themselves will not be treated specially;
a layer can be thought of simply as the collection of arcs sharing
the same type information.

\subsection{Multiple arcs and labels}
\label{sec:multiple}

It is often the case that a given stretch of speech
has multiple possible labels.  For example, the
region of speech corresponding to a monosyllabic
word is both a syllable and a word, and in some
cases it may also be a complete utterance.
The combination of two independent annotations into
a single annotation (through set union) may also result
in two labels covering the same extent.

\begin{figure}[t]
\centerline{\epsfig{figure=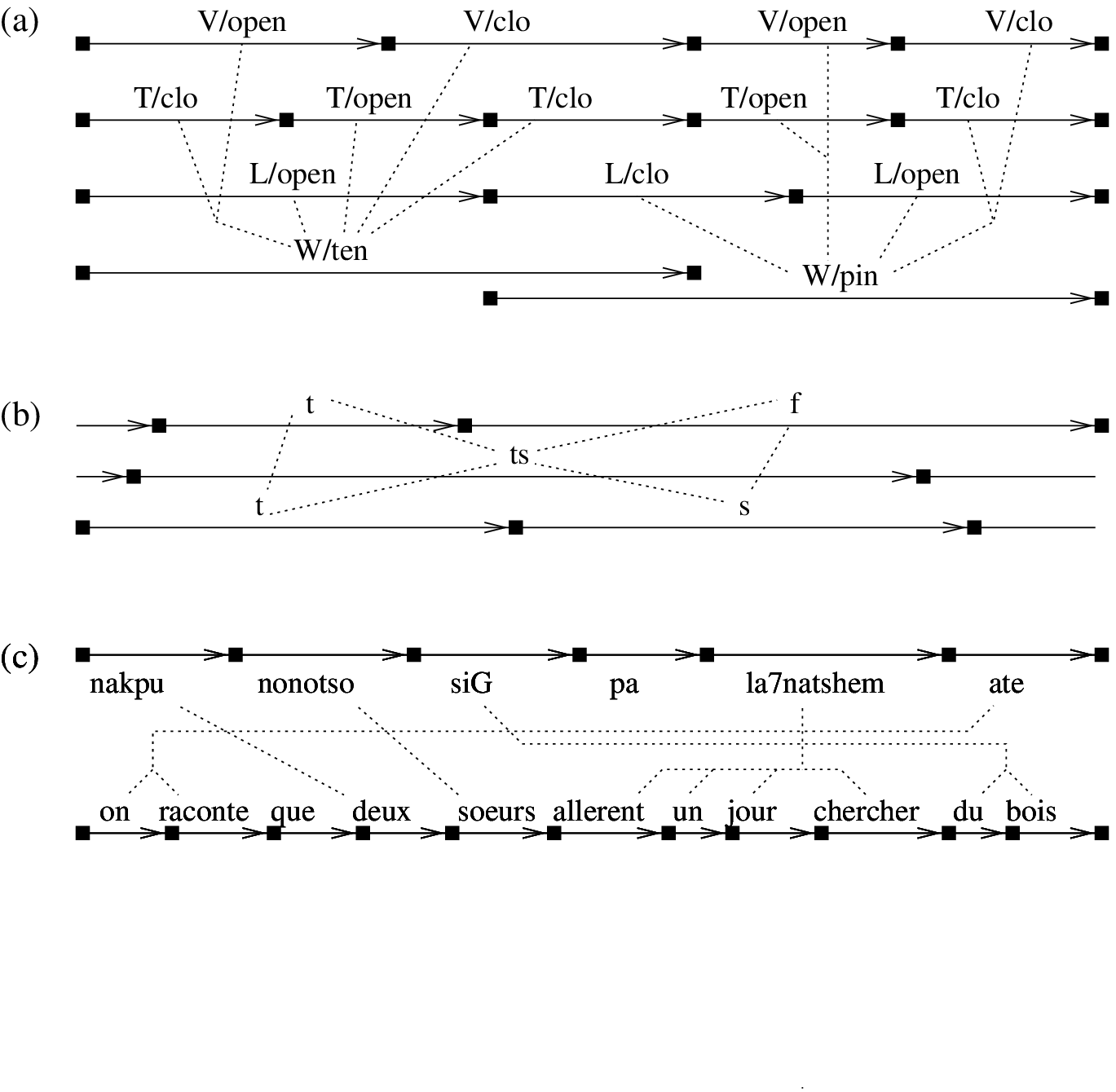,width=.75\linewidth}}
\vspace*{-10ex}
\caption{Annotation Graphs Enriched with Inter-Arc Linkages}\label{links}
\vspace*{2ex}\hrule
\end{figure}

In the general case, a label could be a (typed)
attribute-value matrix, possibly incorporating
nested structure, list- and set-valued attributes,
and even disjunction.
However, our hypothesis is that typed labels (with
atomic types and labels) are sufficient.  Multiple labels spanning the
same material reside on their own arcs.  Their
endpoints can be varied independently (see \S\ref{sec:scores}),
and the combining and projection of
annotations does not require the merging and splitting of arcs.
An apparent weakness of this conception is that
we have no way of individuating
arcs, and it is not possible for arcs to
reference each other.  However, there are cases when such
links between arcs are necessary.
Three examples are displayed
in Figure~\ref{links}; we discuss each in turn.

Recall from \S\ref{sec:scores} that an annotation graph can
contain several independent streams of information,
where no nodes are shared between the streams.  The
temporal extents of the gestures in the different streams
are almost entirely asynchronous; any equivalences are likely to be
coincidences.  However, these gestures may still have determinate abstract
connections to elements of a phonological analysis. Thus a velar
opening and closing gesture may be associated with a particular nasal
feature, or with a set of nasal features, or with the sequence of
changes from non-nasal to nasal and back again.  But these associations
cannot usually be established purely as a matter of temporal
coincidence, since the phonological features involved are bundled
together into other units (segments or syllables or whatever)
containing other features that connect to other gestures whose
temporal extents are all different.  The rules
of coordination for such gestures involve phase
relations and physical spreading which are completely
arbitrary from the perspective of the representational framework.

A simplified example of the arbitrary relationship between
the gestures comprising a word is illustrated in
Figure~\ref{links}a.  We have the familiar annotation
structure (taken from Figure~\ref{tenpin3}), enriched with
information about which words license which gestures.  The
words are shown as overlapping, although this is not crucially
required.  In the general case, the relationship between words
and their gestures is not predictable from the temporal structure
and the type structure alone.

The example in Figure~\ref{links}b shows a situation
where we have multiple independent
transcriptions of the same data.  In this case, the purpose
is to compare the performance of different transcribers
on identical material.
Although the intervals do not line up exactly,
an obvious correspondence exists between the
labels and it should be possible to navigate between
corresponding labels, even though their precise temporal
relationship is somewhat arbitrary.
Observe that the cross references do not have equivalent
status here; the relationship between \smtt{ts} and \smtt{t} is not
the same as that between \smtt{s} and \smtt{f}.

The final example, Figure~\ref{links}c, shows an annotation
graph based on the Hayu example from Figure~\ref{archivage}.
We would like to be able to navigate between words of a phrasal
translation and the corresponding Hayu words.  This would be
useful, for example, to study the various ways in which a particular
Hayu word is idiomatically translated.  Note that the temporal
relationship between linked elements is much more chaotic here,
and that there are examples of one-to-many and many-to-many
mappings.  The words being mapped do not even need
to be contiguous subsequences.

One obvious way to address these three examples is to permit arc
labels to carry cross-references to other arc labels. The semantics of
such cross-references might be left up to the individual case.  This
requires at least some arcs to be individuated (as all nodes are
already).
While it would be a simple matter to individuate arcs
(c.f.\ \S\ref{sec:extensions}),
this step is not forced on us.  There is another
approach that stays more nearly within the confines of the existing
formalism.  In this approach, we treat all of the cases described
above in terms of equivalence classes.  One way to formalize a set of
equivalence classes is as an ordered pair: class-type:identifier.  But
this is just our label notation all over again -- the only news is
that for label types interpreted as denoting equivalence classes,
different labels with the same identifier are viewed as forming an
equivalence class.  Another way to put this is that two (or more)
labels are connected not by referencing one another, but by jointly
referencing a particular equivalence class.

\begin{figure}
\centerline{\epsfig{figure=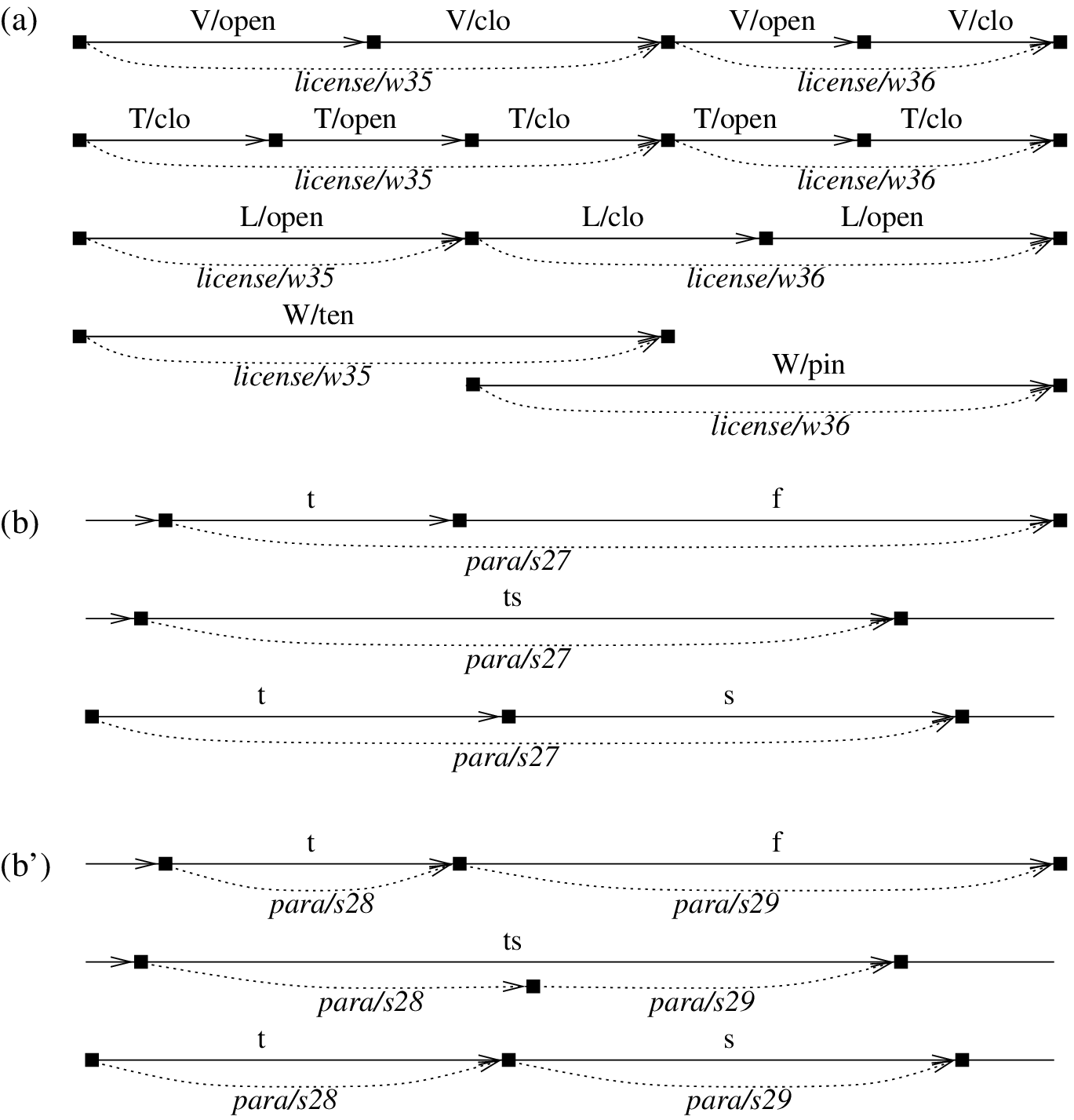,width=.75\linewidth}}
\caption{Inter-Arc Linkages for Parallel Transcriptions}\label{links2}
\vspace*{2ex}\hrule
\end{figure}

In the general case, we have $n$ partially independent
strands, where the material to be associated comes from
some subset of the strands.  Within a given strand,
zero, one or more arcs may participate in a given association,
and the arcs are not necessarily contiguous.  For the
gestural score in Figure~\ref{links}a we augment each
arc with a second arc having the same span.  These additional
arcs all carry the type \smtt{license/} and the unique labels
(say) \smtt{w35} and \smtt{w36}, depending on which word they
belong to.  The word arcs are also supplemented:
\smtt{W/ten} with \smtt{license/w35} and
\smtt{W/pin} with \smtt{license/w36}.  See Figure~\ref{links2}a.
Now we can easily navigate
around the set of gestures licensed by a word regardless of their
temporal extent.  We can use the type information on the
existing labels in situations where we care about the directionality
of the association.

This approach can be applied to the other cases, with some
further qualifications.  For Figure~\ref{links}b, there is
more than one option, as shown in Figure~\ref{links2}b,b'.
In the first option, we have a single cross-reference,
while in the second option, we have two cross-references.
We could combine both of these into a single graph containing
three cross-references.

The translation case of Figure~\ref{links}c can be treated
in the same way.  If the phrasal translation of a word is
a continuous stretch, it could be covered by multiple arcs
(one for each existing arc), or it could be covered by just
a single arc.  If the phrasal translation of
a word is not a contiguous stretch, we may be forced to attach
more than one diacritic arc with a given label.  We do not anticipate
any adverse consequences of such a move.  Incidentally,
note that this linked multiple stream representation is
employed in an actual machine translation system \cite{Brown90}.

Observe that this construction involves assigning intervals
(node-pairs) rather than arcs to equivalence classes.
In cases where there are multiple independent cross references,
it is conceivable that we might have distinct equivalence classes
involving different arcs which span the same two nodes.  So long
as these arcs are distinguished by their types we do not foresee
a problem.

This section has described three situations where potentially
complex relationships between arc labels are required.
However, we have demonstrated that the existing formalism
is sufficiently expressive to encompass such relationships,
and so we are able to preserve the simplicity of the model.
Despite this simplicity, there is one way in which the
approach may seem profligate.  There are
no less than three ways for a pair of arcs to be `associated':
temporal overlap, hierarchy, and equivalence-class linkages.
Interestingly, this three-way possibility exactly mirrors the
three ways that association is treated in the phonological literature.
There, association is first and foremost a graphical notion.
{}From context it is usually possible to tell whether the line
drawn between two items indicates temporal overlap, a hierarchical
relationship, or some more abstract, logical relationship
\cite{BirdKlein90,BirdLadd91,Bird95}.
We have shown how all three uses are attested in the realm of
linguistic annotation.  The fact that the three conceptions of
association are distinct and attested is sufficient cause for
us to include all three in the formalism, notwithstanding the
fact that we get them for free.

\subsection{Associations between annotations and files}
\label{sec:associations}

An `annotated corpus' is a set of annotation graphs and an associated
body of time series data.  The time series might comprise one or more
audio tracks, one or more video streams, one or more streams of
physiological data of various types, and so forth. The data might be
sampled at a fixed rate, or might consist of pairs of times and
values, for irregularly spaced times. Different streams will typically
have quite different sampling rates. Some streams might be defined
only intermittently, as in the case of a continuous audio recording
with intermittent physiological or imaging data. This is not an
imagined list of conceptually possible types of data -- we are familiar
with corpora with all of the properties cited.

The time series data will be packaged into a set of one or more files.
Depending on the application, these files may have some more or less
complex internal structure, with headers or other associated
information about type, layout and provenance of the data. These
headers may correspond to some documented open standard, or they may
be embedded in a proprietary system.

The one thing that ties all of the time series data together is a
shared time base. To use these arbitrarily diverse data streams, we
need to be able to line them up time-wise. This shared time base is also
the only pervasive and systematic connection such data is likely to have
with annotations of the type we are discussing in this paper.

It is not appropriate for an annotation framework to try to encompass
the syntax and semantics of all existing time series file
formats. They are simply too diverse and too far from being
stable. However, we do need to be able to specify what time series
data we are annotating, and how our annotations align with it, in a
way that is clear and flexible.

An ambitious approach would be to specify a new universal framework
for the representation of time series data, with a coherent and
well-defined semantics, and to insist that all annotated time series
data should be translated into this framework. After all, we are doing
the analogous thing for linguistic annotations: proposing a new,
allegedly universal framework into which we argue that all annotations
can be translated. Such an effort for all time series data, whether or
not it is a reasonable thing to do, is far outside the scope of what
we are attempting here.

A much simpler and less ambitious way to connect annotation graphs to
their associated time series is to introduce arcs that reference
particular time-series files, or temporally contiguous sub-parts of
such files. Each such arc specifies that the cited portion of data in
the cited time-function file lines up with the portion of the
annotation graph specified by the time-marks on its source and sink
nodes. Arbitrary additional information can be provided, such as an
offset relative to the file's intrinsic time base (if any), or a
specification selecting certain dimensions of vector-valued data.
Taking this approach, a single annotation could reference multiple
files -- some parts of an annotation could refer specifically to
a single file, while other parts of an annotation could be
non-specific.  In this way, events that are specific to a channel
(like a particular speaker turn) can be marked as such.
Equally, annotation content for an event which is not specific
to a channel can be stored just once.

These file-related labels, if properly designed and implemented, will
permit an application to recover the time-series data that corresponds
to a given piece of annotation -- at least to the extent that the
annotation is time-marked and that any time-function files have been
specified for the cited subgraph(s). Thus if time-marking is provided
at the speaker-turn level (as is often the case for published
conversational data), then a search for all the instances of a
specified word string will enable us to recover usable references to
all available time-series data for the turn that contains each of
these word strings. The information will be provided in the form of
file names, time references, and perhaps time offsets; it will be the
responsibility of the application (or the user) to resolve these
references. If time-marking has been done at the word level, then the
same query will enable us to recover a more exact set of temporal
references in the same set of files.

Our preference for the moment is to allow the details of how to define
these file-references to fall outside the formalism we are defining
here. It should be clear that there are simple and natural ways to
establish the sorts of linkages that are explicit in existing types of
annotated linguistic database.  After some practical experience,
it may make sense to try to provide a more formal account of references
to external time-series data.

\subsubsection*{Spatial and image-plane references}

We would also like to point out a wider problem for which we do not
have any general solution.  Although it is not our primary focus, we
would like the annotation formalism to be extensible to
spatially-specific annotations of video signals and similar data,
perhaps by enriching the
temporal anchors with spatial and/or image-plane information.  Anthropologists,
conversation analysts,
and sign-language researchers
are already producing annotations that are (at least
conceptually) anchored not only to time spans but also to a particular
spatial or image-plane trajectory through the corresponding series of video frames.

In the case of simple time-series annotations, we are tagging nodes with absolute
time references, perhaps offset by a single constant for
a given recorded signal.
However, if we are annotating a video recording,
the additional anchoring used for annotating video sequences
will mostly not be about absolute space, even with
some arbitrary shift of coordinate origin, but
rather will be coordinates in the image plane. If there
are multiple cameras, then image coordinates for each will differ,
in a way that time marks for multiple simultaneous recordings
do not.

In fact, there are some roughly similar cases in audio annotation,
where an annotation might reference some specific two- or
three-dimensional feature of (for instance) a time-series of
short-time amplitude spectra (i.e.\ a spectrogram), in which case the
quantitative details will depend on the analysis recipe. Our system
allows such references (like any other information) to be encoded in
arc labels, but does not provide any more specific support.

\subsubsection*{Relationship to multimedia standards}

In this context we ought to raise the question of how
annotation graphs relate to various multimedia standards like
the Synchronized Multimedia Integration Language
[\smtt{www.w3.org/TR/REC-smil/}]
and MPEG-4 [\smtt{drogo.cselt.it/mpeg/standards/mpeg-4/mpeg-4.htm}].
Since these provide ways to specify both
temporal and spatial relationships among strings, audio clips, still
pictures, video sequences, and so on, one hopes that they
will offer support for linguistic annotation. It is hard to offer a
confident evaluation, since MPEG-4 is still in development, and SMIL's
future as a standard is unclear. 

With respect to MPEG-4, we reserve judgment until its
characteristics become clearer.
Our preliminary assessment is that SMIL is not useful for purposes of
linguistic annotation, because it is mainly focused on presentational
issues (fonts, colors, screen locations, fades and animations, etc.)
and does not in fact offer any natural ways to encode the sorts of
annotations that we surveyed in the previous section. Thus it is easy
to specify that a certain audio file is to be played while a certain
caption fades in, moves across the screen, and fades out. It is not
(at least straightforwardly) possible to specify that a certain audio
file consists of a certain sequence of conversational turns,
temporally aligned in a certain way, which consist in turn of certain
sequences of words, etc.


\subsection{Node references versus byte offsets}
\label{sec:references}

The Tipster Architecture for linguistic annotation of text \cite{Grishman96}
is based on the concept of a fundamental, immutable textual foundation,
with all annotations expressed in terms of byte offsets into this
text. This is a reasonable solution for cases where the text
is a published given, not subject to revision by annotators.
However, it is not a good solution for speech transcriptions,
which are typically volatile entities, constantly up for revision both
by their original authors and by others.

In the case of speech transcriptions, it is more appropriate to treat
the basic orthographic transcription as just another annotation, no
more formally privileged than a discourse analysis or a translation.
Then we are in a much better position to deal with the common
practical situation, in which an initial orthographic transcription of
speech recordings is repeatedly corrected by independent users, who
may also go on to add new types of annotation of their own, and
sometimes also adopt new formatting conventions to suit their own
display needs. Those who wish to reconcile these independent
corrections, and also combine the independent additional annotations,
face a daunting task. In this case, having annotations reference byte
offsets into transcriptional texts is almost the worst imaginable
solution.

Although nothing will make it trivial to untangle this situation,
we believe our approach comes close.  As we shall see in \S\ref{sec:files},
our use of a flat, unordered file structure incorporating
node identifiers and time references means that
edits are as strictly local as they possibly can be, and
connections among various types of annotation are as durable as they
possibly can be. Some changes are almost completely transparent
(e.g.\ changing the spelling of a name). Many other changes will turn
out not to interact at all with other types of annotation. When there
is an interaction, it is usually the absolute minimum that is
necessary. Therefore, keeping track of what corresponds to what, across
generations of distributed annotation and revision, is as simple as
one can hope to make it.

Therefore we conclude that
Tipster-style byte offsets are an inappropriate choice for use as
references to audio transcriptions, except for cases
where such transcriptions are immutable in principle.

In the other direction, there are several ways to translate Tipster-style
annotations into our terms. The most direct way would be to treat Tipster byte
offsets exactly as analogous to time references -- since the only formal
requirement on our time references is that they can be ordered. This
method has the disadvantage that the underlying text could not be
searched or displayed in the same way that a speech transcription normally could.
A simple solution would be to add an arc for each of the lexical
tokens in the original text, retaining the byte offsets on the corresponding
nodes for translation back into Tipster-architecture terms.

\subsection{What is time?}
\label{sec:offsets}

TIMIT and some other extant databases denominate signal time in sample
numbers (relative to a designated signal file, with a known sampling
rate). Other databases use floating-point numbers, representing time
in seconds relative to some fixed offset, or other representations of
time such as centiseconds or milliseconds. In our formalization of
annotation graphs, the only thing that really matters about time
references is that they define an ordering. However,
for comparability across signal types, time references need to be
intertranslatable.

We feel that time in seconds is generally preferable to sample or
frame counts, simply because it is more general and easier to
translate across signal representations. However, there may be
circumstances in which exact identification of sample or frame numbers
is crucial, and some users may prefer to specify these directly to
avoid any possibility of confusion.  

Technically, sampled data points (such as audio samples or video
frames) may be said to denote time intervals rather than time points,
and the translation between counts and times may therefore become
ambiguous. For instance, suppose we have video data at 30 Hz. Should
we take the 30th video frame (counting from one) to cover the time
period from 29/30 to 1 second or from 29.5/30 to 30.5/30 second?  In
either case, how should the endpoints of the interval be assigned?
Different choices may shift the correspondence between times and frame
numbers slightly.

Also, when we have signals at very different sampling rates, a single
sampling interval in one signal can correspond to a long sequence of
intervals in another signal.
With video at 30 Hz and audio at 44.1 kHz, each video frame
corresponds to 1,470 audio samples. Suppose we have a time reference
of .9833 seconds.  A user might want to know whether this was
created because some event was flagged in the 29th video frame, for
which we take the mean time point to be 29.5/30 seconds, or because some
event was flagged at the 43,365th audio sample, for which we take the
central time point to be 43365.5/44100 seconds.

For reasons like these, some users might want the freedom to
specify references explicitly in terms of sample or frame numbers,
rather than relying on an implicit method of translation to and from
time in seconds.

Several ways to accommodate this within our framework come to mind, but
we prefer to leave this open, as we have no experience with
applications in which this might be an issue. In our initial explorations,
we are simply using time in seconds as the basis.

\section{A Formal Framework}
\label{sec:algebra}

\subsection{Background}

Looking at the practice of speech transcription and annotation
across many existing `communities of practice', we see commonality of
abstract form along with diversity of concrete format.

All annotations of recorded linguistic signals require one unavoidable basic
action: to associate a label, or an ordered sequence of labels, with a
stretch of time in the recording(s). Such annotations also typically
distinguish labels of different types, such as spoken words vs.\ non-speech
noises. Different types of annotation often span different-sized
stretches of recorded time, without necessarily forming a strict
hierarchy: thus a conversation contains (perhaps overlapping)
conversational turns, turns contain (perhaps interrupted) words, and
words contain (perhaps shared) phonetic segments.

A minimal formalization of this basic set of practices is a directed
graph with typed labels on the arcs and optional time references on
the nodes.  We believe that this minimal formalization in fact has
sufficient expressive capacity to encode, in a reasonably intuitive
way, all of the kinds of linguistic annotations in use today.  We also
believe that this minimal formalization has good properties with
respect to creation, maintenance and searching of annotations.

Our strategy is to see how far this simple conception can go,
resisting where possible the temptation to enrich its ontology
of formal devices, or to establish label types with special syntax or
semantics as part of the formalism. See section \S\ref{sec:extensions}
for a perspective on how to introduce formal and substantive extensions into
practical applications.

We maintain that most, if not all, existing annotation formats
can naturally be treated, without loss of generality,
as directed acyclic graphs having
typed labels on (some of) the edges and time-marks on
(some of) the vertices.  We call these `annotation graphs'.
It is important to recognize that translation into
annotation graphs does not magically create compatibility among
systems whose semantics are different.  For instance, there are many
different approaches to transcribing filled pauses in English -- each
will translate easily into an annotation graph framework, but their
semantic incompatibility is not thereby erased.

It is not our intention here to specify annotations at the level of
permissible tags, attributes, and values, as was done by many of the
models surveyed in \S\ref{sec:survey}.  This is an application-specific
issue which does not belong in the formalism.  The need for
this distinction can be brought into sharp focus by analogy with
database systems.  Consider the relationship between the abstract
notion of a relational algebra, the features of a relational database
system, and the characteristics of a particular database.  For
example, the definition of substantive notions like `date' does not
belong in the relational algebra, though there is good reason for a database
system to have a special data type for dates.  Moreover, a
particular database may incorporate all manner of restrictions on
dates and relations among them.  The formalization presented here is
targeted at the most abstract level: we want to get the annotation
formalism right.  We assume that system implementations will add all
kinds of special-case data types (i.e.\ types of labels with
specialized syntax and semantics).  We further assume that particular
databases will want to introduce additional specifications.

Our current strategy -- given the relative lack of experience of the field
in dealing with such matters -- is to start with a general model with
very few special label types, and an open mechanism for allowing users
to impose essentially arbitrary interpretations.
This is how we deal with instants (c.f.~\S\ref{sec:instants}),
associations between annotations and files (c.f.~\S\ref{sec:associations})
and coindexing of arcs (c.f.~\S\ref{sec:multiple}).

\subsection{Annotation graphs}
\label{sec:formalism}

\newtheorem{defn}{Definition}

Let $T$ be a set of types, where each type in $T$
has a (possibly open) set of contentful elements.
The label space $L$ is the union of all these sets.
We write each label as a \smtt{$<$type$>$/$<$content$>$} pair,
allowing the same contentful element to occur in different types.
(So, for example, the phoneme /a/ and the phonetic segment [a]
can be distinguished as \smtt{P/a} vs \smtt{S/a}.)
Annotation graphs are now defined as follows:

\begin{defn}
An \textbf{annotation graph} $G$ over a label set
$L$ and a node set $N$ is a set of triples having the form
$\left<n_1, l, n_2\right>$, $l\in L$, $n_1, n_2\in N$, which
satisfies the following conditions:

\begin{enumerate}
\item
\(
  \left<
    N,
    \left\{
      \left<n_1,n_2\right> \mid
      \left<n_1,l,n_2\right> \in G
    \right\}
  \right>
\)
is a directed acyclic graph.

\item $\tau: N \rightharpoonup \Re$
is an order-preserving map
assigning times to some of the nodes.
\end{enumerate}
\end{defn}

There is no requirement that annotation graphs be connected
or rooted, or that they cover the whole time course of the
linguistic signal they describe.
The set of annotation graphs is
closed under union, intersection and relative complement.

For convenience, we shall refer to nodes which have a time
reference (i.e.\ $\dom(\tau)$) as {\it anchored nodes}.
It will also be useful to talk about annotation graphs which
are minimally anchored, in the sense defined below:

\begin{defn}
An \textbf{anchored annotation graph} $G$ over a label set
$L$ and a node set $N$ is an annotation graph satisfying two
additional conditions:

\begin{enumerate}
\item
If $n\in N$ is such that
$\left<n, l, n'\right> \not\in G$ for any $l \in L$, $n' \in N$,
then $\tau: n \mapsto r \in \Re$;

\item
If $n\in N$ is such that
$\left<n', l, n\right> \not\in G$ for any $l \in L$, $n' \in N$,
then $\tau: n \mapsto r \in \Re$.
\end{enumerate}
\end{defn}

Anchored annotation graphs have no dangling arcs (or chains) leading
to an indeterminate time point.  It follows from this definition
that, for any unanchored node, we can reach an anchored node by
following a chain of arcs.  In fact every path from an unanchored
node will finally take us to an anchored node.  Likewise, an
unanchored node can be reached from an anchored node.  A key property
of anchored annotation graphs is that we are guaranteed to have
some information about the temporal locus of every node.
This property will be made explicit in \S\ref{sec:time-local}.
An examination of the annotation graphs in
\S\ref{sec:survey} will reveal that they are all anchored annotation
graphs.

Note that the set of anchored annotation graphs is closed under
union, but not under intersection or relative complement.

We can also define a {\it totally-anchored annotation graph}
as one in which $\tau$ is a total function.  The annotation
graphs in Figures~\ref{timit}, \ref{partitur}, \ref{chat2} and \ref{emu1}
are all totally-anchored.

Equipped with this three-element hierarchy, we will insist that
the annotation graphs that are the primary objects in linguistic
databases are anchored annotation graphs.
For the sake of a clean algebraic semantics for the query language,
we will permit queries and the results of queries to be (sets of)
arbitrary annotation graphs.


\subsection{Relations on nodes and arcs}

The following definition lets us talk about two kinds
of precedence relation on nodes in the graph structure.
The first kind respects the graph structure (ignoring the
time references), and is called
structure precedence, or simply {\it s-precedence}.  The second
kind respects the temporal structure (ignoring the graph
structure), and is called temporal precedence, or simply {\it t-precedence}.

\begin{defn}
A node $n_1$ \textbf{s-precedes} a node $n_2$,
written $n_1 <_s n_2$, if there is a chain from $n_1$ to $n_2$.
A node $n_1$ \textbf{t-precedes} a node $n_2$,
written $n_1 <_t n_2$, if $\tau(n_1) < \tau(n_2)$.
\end{defn}

Observe that both these relations are transitive.
There is a more general notion of precedence which
mixes both relations.  For example, we can infer that
node $n_1$ precedes node $n_2$ if we can use a mixture
of structural and temporal information to get from $n_1$
to $n_2$.  This idea is formalized in the next definition.

\begin{defn}
\textbf{Precedence} is a binary relation on nodes, written $<$,
which is the transitive closure of the union of the s-precedes and the
t-precedes relations.
\end{defn}

Armed with these definitions we can now define some useful
inclusion relations on arcs.
The first kind of inclusion respects the graph
structure, so it is called {\it s-inclusion}.
The second kind, {\it t-inclusion}, respects the temporal structure.
\begin{defn}
An arc $p = \left<n_1, n_4\right>$ \textbf{s-includes}
an arc $q = \left<n_2, n_3\right>$,
written $p \supset_s q$, if $n_1 <_s n_2$ and $n_3 <_s n_4$.
$p$ \textbf{t-includes} $q$, written $p \supset_t q$, if 
$n_1 <_t n_2$ and $n_3 <_t n_4$.
\end{defn}

As with node precedence, we define a general notion
of inclusion which generalizes over these two types:

\begin{defn}
\textbf{Inclusion} is a binary relation on arcs, written $\supset$,
which is the transitive closure of the union of the s-inclusion and the
t-inclusion relations.
\end{defn}

Note that all three inclusion relations are
reflexive and transitive.  We assume the existence of
non-strict precedence and inclusion relations, defined
in the obvious way.

\subsection{Visualization}
\label{sec:visualization}

It is convenient to have a variety of ways of visualizing
annotation graphs.  Most of the systems we surveyed in
\S\ref{sec:survey} come with visualization components,
whether tree-based, extent-based, or some combination of
these.  We would endorse the use of any descriptively adequate
visual notation in concert with the annotation graph formalism,
so long as the notation can be endowed with an explicit
formal semantics in terms of annotation graphs.
Note, however, that not all such visual notations can represent
everything an annotation graph contains, so we still need one or more
general-purpose visualizations for annotation graphs.

The primary visualization chosen for annotation graphs in this paper
uses networks of nodes and arcs to make the point that the
mathematical objects we are dealing with are graphs.  In most practical
situations, this mode of visualization is cumbersome to the point of
being useless. Visualization techniques should be optimized for each
type of data and for each application, but there are some general
techniques that can be cited.

Observe that
the direction of time-flow can be inferred from the left-to-right
layout of annotation graphs, and so the arrow-heads are
redundant.  For simple connected sequences (e.g.\ of words) the linear structure
of nodes and arcs is not especially informative; it is better
to write the labels in sequence and omit the graph
structure.  The ubiquitous node identifiers should not be displayed
unless there is accompanying text that refers to specific
nodes.  Label types can be effectively distinguished
with colors, typefaces or vertical position.  We will usually need to break
an annotation graph into chunks which can be presented line-by-line
(much like interlinear text) in order to fit on a screen or a page.

The applicability of these techniques depends on the fact that
annotation graphs have a number of properties that
do not follow automatically from a graphical notation.
In other words, many directed acyclic graphs are not
well-formed annotation graphs.

\begin{figure}
\centerline{\epsfig{figure=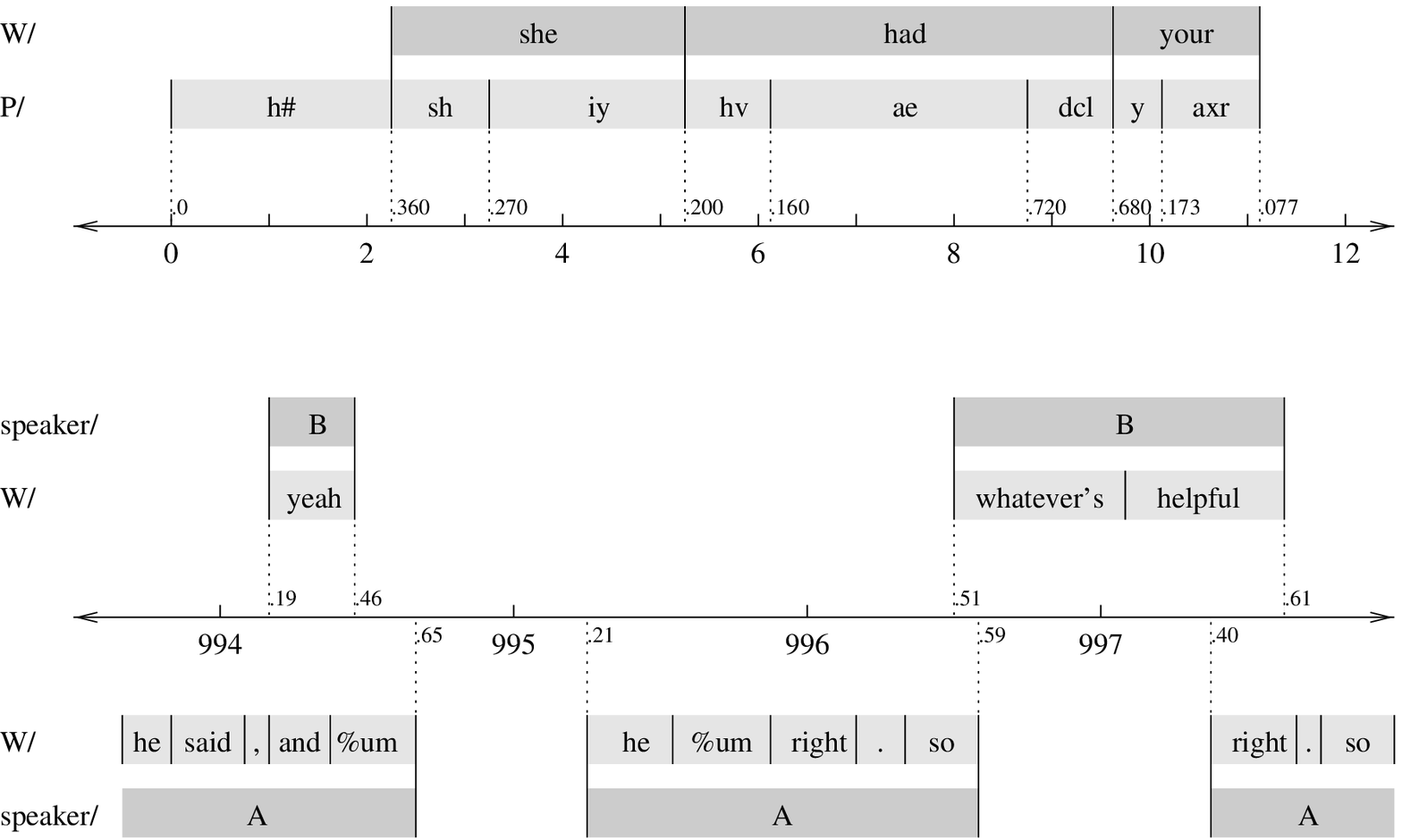,width=.9\linewidth}}
\caption{Visualizations for the TIMIT and LDC Telephone Speech Examples}\label{visual}
\vspace*{2ex}\hrule
\end{figure}

Two properties are of particular interest here.  First, as
noted in \S\ref{sec:formalism}, all the annotation graphs we
have surveyed are actually anchored annotation graphs.
This means that every arc lies on a path of arcs that
is bounded at both ends by time references.  So, even when most
nodes lack a time reference, we can still associate
such chains with an interval of time.
A second property, more contingent but equally convenient, is that
annotation graphs appear to be `rightward planar',
i.e.\ they can be drawn in such a way that no arcs cross
and each arc is monotonically increasing in the rightwards direction
(c.f.\ the definition of upward planarity in \cite{Battista94}).
These properties are put to good use in Figure~\ref{visual},
which employs a score notation
(c.f.\ \cite{Browman89,Cassidy96,Ehlich92,Neidle98}).

The conventions employed by these diagrams are as follows.
An arc is represented by a shaded rectangle, where the shading
(or color, if available) represents the type information.
Where possible, arcs having the same type are displayed on
the same level.  Arcs are labeled, but the type information is
omitted.  Inter-arc linkages (see \S\ref{sec:multiple})
are represented using coindexing.
The ends of arcs are represented using short vertical lines
having the same width as the rectangles.  These may be omitted
if the tokenization of a string is predictable.
If two arcs are
incident on the same node but their corresponding rectangles
appear on different levels of the diagram, then the relevant
endpoints are connected by a solid line.  For ease
of external reference, these lines can be decorated with a node
identifier.
Anchored nodes are connected to the timeline with dotted lines.
The point of intersection is labeled with a time reference.
If necessary, multiple timelines may be used.
Nodes sharing a time reference are connected with a dotted line.
In order to fit on a page, these diagrams may be cut at any point,
with any partial rectangles labeled on both parts.

Unlike some other conceivable visualizations (such as the tree diagrams
and autosegmental diagrams used by Festival and Emu),
this scheme emphasizes the fact that each
component of an annotation has temporal extent.  The scheme neatly
handles the cases where temporal information is partial.

\subsection{File Encodings}
\label{sec:files}

As stated at the outset, we believe that the standardization
of file formats is a secondary issue.  The identification of
a common conceptual framework underlying all work in this
area is an earlier milestone along any path to standardization
of formats and tools.
That said, we believe that file formats should be transparent
encodings of the annotation structure.

The flattest data structure we can imagine for an annotation
graph is a set of 3-tuples, one per arc,
consisting of a node identifier, a label, and another node identifier.
This data structure has
a transparent relationship to our definition of annotation graphs,
and we shall refer to it as the `basic encoding'.
Node identifiers are supplemented with time values, where available,
and are wrapped with angle brackets.
A file encoding for the UTF example (Figure~\ref{utf}) is
given below.

\begin{bv}
<21/3291.29> speaker/Gloria-Allred <25/2439.82>
<13/2391.11> W/country <14/2391.60>
<11/2348.81> spkrtype/male <14/2391.60>
<21/3291.29> spkrtype/female <25/2439.82>
<22/> W/i <23/2391.60>
<23/2391.60> W/think <24/>
<11/2348.81> speaker/Roger-Hedgecock <14/2391.60>
<12/> W/this <13/2391.11>
<21/3291.29> W/well <22/>
\end{bv}

We make no ordering requirement, thus any reordering of these lines is
taken to be equivalent.  Equally, any subset of the tuples comprising
an annotation graph (perhaps determined by matching a `grep' like
pattern) is a well-formed annotation graph.
Accordingly, a basic query operation on an annotation graph can be
viewed as asking for subgraphs that satisfy some predicate,
and each such subgraph will itself be an annotation graph.
Any union of the tuples comprising annotation graphs is a well-formed
annotation graph, and this can be implemented by simple concatenation
of the tuples (ignoring any repeats).

This format obviously encodes redundant information,
in that nodes and their time references may be mentioned more than once.
However, we believe this is a small price to pay for
having a maximally simple file structure.

Let us consider the implications of various kinds of annotation updates
for the file encoding.
The addition of new nodes and arcs simply involves concatenation
to the basic encoding (recall that the basic encoding is an unordered
list of arcs).  The same goes for the addition of new arcs
between existing nodes.
For the user adding new annotation data to an existing
read-only corpus -- a widespread mode of operation --
the new data can reside in one or more separate files,
to be concatenated at load time.
The insertion and modification
of labels for existing arcs involves changing one line of
the basic encoding.

Adding, changing or deleting a time reference may involve non-local
change to the basic encoding of an annotation.  This can be done in either of
two ways: a linear scan through the basic encoding, searching
for all instances of the node identifier; or indexing into the
basic encoding using the time-local index to find the relevant
lines.  Of course, the
time reference could be localized in the basic encoding by
having a separate node set, referenced by the arc set.
This would permit the time reference of a node
to be stored just once.  However, we prefer to keep the basic
encoding as simple as possible.

Maintaining consistency of the temporal and hierarchical structure
of an annotation under updates requires further consideration.
In the worst case, an entire annotation structure would have to
be validated after each update.  To the extent that information
can be localized, it is to be expected that incremental validation
will be possible.  This might apply after each and every update,
or after a collection of updates in case there is a sequence of elementary
updates which unavoidably takes us to an invalid structure along
the way to a final, valid structure.

Our approach to the file encoding has some interesting implications
in the area of so-called `standoff markup' \cite{Thompson97}.  Under our
proposed scheme, a readonly
file containing a reference annotation can be concatenated with
a file containing additional annotation material.  In order for
the new material to be linked to the existing material, it simply
has to reuse the same node identifiers
and/or have nodes anchored to the same time base.
Annotation deltas can employ a `diff' method operating
at the level of individual arcs.  Since the file contains
one line per arc and since arcs are unordered, no context needs
to be specified other than the line which is being replaced or modified.
A consequence of our approach is that all speech annotation
(in the broad sense) can be construed as `standoff' description.

\section{Indexing}
\label{sec:indexing}

Corpora of annotated texts and recorded signals
may range in size from a few thousand words up into
the billions.  The data may be in the form of a monolithic
file, or it may be cut up into word-size pieces, or
anything in between.  The annotation might be dense
as in phonetic markup or sparse as in discourse markup,
and the information may be uniformly or sporadically distributed
through the data.

At present, the annotational components of most speech
databases are still relatively small objects.
Only the largest annotations would cover a whole hour of
speech (or 12,000 words at 200 words per minute), and
even then, a dense annotation of this much material would only
occupy a few hundred kilobytes.  In most cases, serial
search of such annotations will suffice.
Ultimately, however, it will be necessary to devise
indexing schemes; these will necessarily be application-specific,
depending on the nature of the corpus and of the
queries to be expressed.  The indexing method is not a
property of the query language but a way to make certain
kinds of query run efficiently.  For large corpora, certain
kinds of query might be essentially useless without such indexing.

At the level of individual arc labels,
we envision three simple indexes, corresponding to the
three obvious dimensions of an annotation graph:
a time-local index, a type-local index and a
hierarchy-local index.  These are discussed below.
More sophisticated indexing schemes could surely be devised,
for instance to support proximity search on node labels.
We also assume the existence of an index for node identifiers;
a simple approach would be to sort the lines of the annotation file
with respect to an ordering on the node identifiers.
Note that, since we wish to index linguistic databases, and
not queries or query results, the indexes will assume that
annotation graphs are anchored.

\subsection{A time-local index}
\label{sec:time-local}

We index the annotation graph in terms of the intervals it employs.
Let $r_i \in R$ be the sequence of time references used by the
annotation.  We form the intervals $[r_i, r_{i+1})$.
Next, we assign each arc to a contiguous set of these
intervals.

Suppose that an arc is incident on nodes which are
anchored to time points
$r_p$ and $r_q$, where $r_p < r_q$.  Then we assign the
arc to the following set of intervals:
\(
\left\{
[r_p, r_{p+1}),
[r_{p+1}, r_{p+2}), \ldots,
[r_{q-1}, r_q)
\right\}
\)
Now we generalize this construction to
work when a time reference is missing from
either or both of the nodes.  First we define the
{\it greatest lower bound (glb)} and the {\it least upper bound (lub)}
of an arc.

\begin{defn}
Let $a = \left<n_1, l, n_2\right>$ be an arc.
\textbf{$\glb(a)$} is the greatest time reference $r \in R$ such that
there is some node $n$ with $\tau(n) = r$ and $n <_s n_1$.
\textbf{$\lub(a)$} is the least time reference $r \in R$ such that
there is some node $n$ with $\tau(n) = r$ and $n_2 <_s n$.
\end{defn}

According to this definition, the {\it glb} of an arc
is the time mark of the `greatest' anchored node from which the arc is
reachable.  Similarly, the {\it lub} of an arc is the time mark of
the `least' anchored node reachable from that arc.
If $a = \left<n_1, l, n_2\right>$ is anchored at both ends then
$\glb(a) = n_1$ and $\lub(a) = n_2$.
The {\it glb} and {\it lub} are guaranteed to exist for
anchored annotation graphs (but not for annotation graphs in general).
The {\it glb} and {\it lub} are guaranteed to be unique since
$R$ is a total ordering.
We can take the {\em potential} temporal span of an arc $a$
to be $[\glb(a), \lub(a))$.  We then assign the arc to a
set of intervals as before.
Below we give an example time-local index for the UTF annotation
from Figure~\ref{utf}.

\begin{bv}
2348.81  2391.11  <12/> W/this <13/2391.11>
                  <11/2348.81> speaker/Roger-Hedgecock <14/2391.60>
                  <11/2348.81> spkrtype/male <14/2391.60>
2391.11  2391.29  <13/2391.11> W/country <14/2391.60>
                  <11/2348.81> speaker/Roger-Hedgecock <14/2391.60>
                  <11/2348.81> spkrtype/male <14/2391.60>
2391.29  2391.60  <13/2391.11> W/country <14/2391.60>
                  <22/> W/i <23/2391.60>
                  <21/3291.29> W/well <22/>
                  <21/3291.29> speaker/Gloria-Allred <25/2439.82>
                  <11/2348.81> speaker/Roger-Hedgecock <14/2391.60>
                  <21/3291.29> spkrtype/female <25/2439.82>
                  <11/2348.81> spkrtype/male <14/2391.60>
2391.60  2439.82  <21/3291.29> speaker/Gloria-Allred <25/2439.82>
                  <21/3291.29> spkrtype/female <25/2439.82>
                  <23/2391.60> W/think <24/>
\end{bv}

The index is built on a sequence of
four temporal intervals which are derived from the
five time references used in Figure~\ref{utf}.
Observe that the right hand side of the index is made
up of fully-fledged arcs (sorted lexicographically),
rather than references to arcs.
Using the longer, fully-fledged arcs
has two benefits.  First, it localizes the arc
information on disk for fast access.  Second, the right
hand side is a well-formed annotation graph which can
be directly processed by the same tools used by the rest
of any implementation, or used as a citation.

This time-local index can be used for computing general
overlap and inclusion relations.
To find all arcs overlapping a given arc $p$, we iterate
through the list of time-intervals comprising $p$ and collect up
the arcs found in the time-local index for each such interval.
Additional checks can be performed to see if a candidate arc
is `s-overlapped' or `t-overlapped'.
This process, or parts of it, could be done offline.

To find all arcs included in a given arc $p$, we can find the
overlapping arcs and perform the obvious tests for s-inclusion
or t-inclusion.  Again, this process could be done offline.

An interesting property of the time-local index is that it
is well-behaved when time information is partial.

\subsection{Type-local indexes}
\label{type-local}

Continuing in the same vein as
the time-local index we propose a set of self-indexing structures
for the types -- one for each type.
The arcs of an annotation graph are then
partitioned into types.  The index for each type
is a list of arcs, sorted as follows (c.f.\ \cite{Sacks-Davis97}):

\begin{enumerate}
\item of two arcs, the one bearing the lexicographically
earlier label appears first;

\item if two arcs share the same label, the one having
the least {\it glb} appears first;

\item if two arcs share the same label and have the same
{\it glb}, then the one with the larger {\it lub} appears first.
\end{enumerate}

\begin{bv}
W        country          <13/2391.11> W/country <14/2391.60>
         i                <22/> W/i <23/2391.60>
         think            <23/2391.60> W/think <24/>
         this             <12/> W/this <13/2391.11>
         well             <21/3291.29> W/well <22/>
speaker  Gloria-Allred    <21/3291.29> speaker/Gloria-Allred <25/2439.82>
         Roger-Hedgecock  <11/2348.81> speaker/Roger-Hedgecock <14/2391.60>
spkrtype female           <21/3291.29> spkrtype/female <25/2439.82>
         male             <11/2348.81> spkrtype/male <14/2391.60>
\end{bv}

\subsection{A hierarchy-local index}
\label{sec:hierarchy-local}

Annotations also need to be indexed with respect to their
implicit hierarchical structure (c.f.~\S\ref{sec:hierarchy}).
Recall that we have two kinds of inclusion relation,
s-inclusion (respecting graph structure) and t-inclusion
(respecting temporal structure).  We refine these relations
to be sensitive to an ordering on our set of types $T$.
This ordering has been left external to the formalism, since
it does not fit easily into the flat structure described
in \S\ref{sec:files}.
We assume the existence of a function $\type(p)$ returning the
type of an arc~$p$.

\begin{defn}
An arc $p$ \textbf{s-dominates} an arc $q$, written $n_1 \triangleright_s n_2$,
if $\type(p) > \type(q)$ and $p \supseteq_s q$.
An arc $p$ \textbf{t-dominates} an arc $q$, written $n_1 \triangleright_t n_2$,
if $\type(p) > \type(q)$ and $p \supseteq_t q$.
\end{defn}

Again, we can define a dominance relation which is neutral between
these two, as follows:

\begin{defn}
An arc $p$ \textbf{dominates} an arc $q$, written $n_1 \triangleright n_2$,
if $\type(p) > \type(q)$ and $p \supseteq q$.
\end{defn}

In our current conception, s-dominance will be the most useful.
(The three kinds of dominance were included for generality and
consistency with the preceding discussion.)

We now illustrate an index for s-dominance.  Suppose the ordering on types is:
\smtt{speaker/} $>$ \smtt{W/} and \smtt{spkrtype/} $>$ \smtt{W/}.
We could index the UTF example as follows, ordering the arcs using the
method described in \S\ref{type-local}, and using indentation to
distinguish the dominating arcs from the dominated arcs.

\begin{bv}
<11/2348.81> speaker/Roger-Hedgecock <14/2391.60>
<11/2348.81> spkrtype/male <14/2391.60>
        <21/3291.29> W/well <22/>
        <22/> W/i <23/2391.60>
        <23/2391.60> W/think <24/>
<21/3291.29> speaker/Gloria-Allred <25/2439.82>
<21/3291.29> spkrtype/female <25/2439.82>
        <12/> W/this <13/2391.11>
        <13/2391.11> W/country <14/2391.60>
\end{bv}

This concludes the discussion of proposed indexes.
We have been deliberately schematic, aiming to demonstrate
a range of possibilities which can be refined and extended later.
Note that the various indexing schemes described above just work
for a single annotation.  We would need to enrich the node-id and
time reference information in order for this to work for whole
databases of annotations (see \S\ref{sec:extensions}).
It could then be generalized further,
permitting search across multiple databases --
e.g.\ to find all instances of a particular word in both
the Switchboard and CallHome English databases (c.f.~\S\ref{sec:callhome}).

Many details about indexes could be
application specific.  Under the approach described
here, we can have several copies of an annotation where each
is self-indexing in a way that localizes different kinds of
information.  A different approach would be to provide three
categories of iterators, each of which takes an arc and returns
the `next' arc with respect to the temporal, sortal and hierarchical
structure of an annotation.
It would be the task of any implementation to make sure that
the basic encoding is consistent with itself, and that the
conglomerate structure (basic encoding plus indexes) is
consistent.

More broadly, the
design of an application-specific indexing scheme will have to
consider what kinds of sequences or
connections among tokens are indexed.  In general, the
indexing method should be based on the same elementary
structures from which queries are constructed.
Indices will specify where particular elementary annotation graphs
are to be found, so a complex search expression
can be limited to those regions
for which these graphs are necessary parts.

\section{Conclusions and Future Work}
\label{sec:conclusion}

\subsection{Evaluation criteria}

There are many existing approaches to linguistic annotation, and many
options for future approaches.  Any evaluation of proposed frameworks,
including ours, depends on the costs and benefits incurred in a range
of expected applications. Our explorations have presupposed a
particular set of ideas about applications, and therefore a particular
set of goals.  We think that these ideas are widely shared, but it
seems useful to make them explicit.

Here we are using `framework' as a neutral term to encompass both the
definition of the logical structure of annotations, as discussed in this
paper, as well as various further specifications of e.g.\ annotation
conventions and file formats.

\begin{description}
\item[Generality, specificity, simplicity]\hfil\\

Annotations should be publishable (and will often be published), and
thus should be mutually intelligible across laboratories, disciplines,
computer systems, and the passage of time.

Therefore, an annotation framework should be sufficiently expressive
to encompass all commonly used kinds of linguistic annotation,
including sensible variants and extensions.  It should be capable of
managing a variety of (partial) information about labels, timing,
and hierarchy.

The framework should also be formally well-defined, and as simple as
possible, so that researchers can easily build special-purpose tools
for unforeseen applications as well as current ones, using future
technology as well as current technology.

\item[Searchability and browsability]\hfil\\

Automatic extraction of information from large annotation databases,
both for scientific research and for technological development, is a key
application.

Therefore, annotations should be conveniently and efficiently searchable,
regardless of their size and content. It should be possible to search
across annotations of different material produced by different groups
at different times -- if the content permits it -- without having to
write special programs. Partial annotations should
be searchable in the same way as complete ones.

This implies that there should be an efficient algebraic query
formalism, whereby complex queries can be composed out of well-defined
combinations of simple ones, and that the result of querying a set
of annotations should be just another set of annotations.

This also implies that (for simple queries) there should be
efficient indexing schemes, providing near constant-time access into
arbitrarily large annotation databases.

The framework should also support easy `projection' of natural
sub-parts or dimensions of annotations, both for searching and for
display purposes. Thus a user might want to browse a complex
multidimensional annotation database -- or the results of a
preliminary search on one -- as if it contained only an orthographic
transcription.

\item[Maintainability and durability]\hfil\\

Large-scale annotations are both expensive to produce and valuable to
retain. However, there are always errors to be fixed, and the
annotation process is in principle open-ended, as new properties can
be annotated, or old ones re-done according to new
principles. Experience suggests that maintenance of linguistic
annotations, especially across distributed edits and additions, can be
a vexing and expensive task.  Therefore, any framework should
facilitate maintenance of coherence in the face of distributed
development and correction of annotations.

Different dimensions of annotation should therefore be orthogonal, in
the sense that changes in one dimension (e.g.\ phonetic transcription)
do not entail any change in others (e.g.\ discourse transcription),
except insofar as the content necessarily overlaps.  Annotations of
temporally separated material should likewise be modular, so that
revisions to one section of an annotation do not entail global
modification.  Queries not affected by corrections or additions
should return the same thing before and after an update.

In order to facilitate use in scientific discourse, it should
be possible to define durable references which remain valid wherever
possible, and produce the same results unless the referenced material
itself has changed.

Note that it is easy enough to define an invertible sequence of
editing operations for any way of representing linguistic annotations
-- e.g.\ by means of Unix `diff' -- but what we need in this case is
also a way to specify the correspondence (wherever it remains defined)
between arbitrary bits of annotation before and after the
edit. Furthermore, we do not want to impose any additional burden on
human editors -- ideally, the work minimally needed to implement a
change should also provide any bookkeeping needed to maintain
correspondences.

\end{description}

How well does our proposal satisfy these criteria? 

We have tried to demonstrate generality, and to provide an adequate
formal foundation, which is also ontologically parsimonious (if not
positively miserly!).

Although we have not defined a query system, we have indicated the
basis on which one can be constructed: (tuple sets constituting)
annotation graphs are closed under union, intersection and
relative complementation; the set of subgraphs of an
annotation graph is simply the power set of its constituent tuples;
simple pattern matching on an annotation graph can be
defined to produce a set of annotation subgraphs; etc.  Obvious sorts of
simple predicates on temporal relations, graphical relations, label
types, and label contents will clearly fit into this framework.

The foundation for maintainability is present: fully orthogonal
annotations (those involving different label types and time points) do
not interact at all, while linked annotations (such as those that
share time points) are linked only to the point that their content
requires. New layers of annotation can be added monotonically, without
any modification whatsoever in the representation of existing
layers. Corrections to existing annotations are as representationally
local as they can be, given their content.

Although we have not provided a recipe for durable citations (or for
maintenance of trees of invertible modifications), the properties just
cited will make it easier to develop practical approaches. In
particular, the relationship between any two stages in the development
or correction of an annotation will always be easy to compute as a set
of basic operations on the tuples
that express an annotation graph. This makes it easy to calculate
just the aspects of a tree or graph of modifications that
are relevant to resolving a particular citation.

\subsection{Future work}
\label{sec:extensions}


\subsubsection*{Interactions with relational data}

Linguistic databases typically include important bodies of information
whose structure has nothing to do with the passage of time in any
particular recording, nor with the sequence of characters in
any particular text. For instance, the Switchboard corpus includes
tables of information about callers (including date of birth, dialect
area, educational level, and sex), conversations (including the
speakers involved, the date, and the assigned topic), and so on. This
side information is usually well expressed as a set of relational tables.

There also may be bodies of relevant information concerning a
language as a whole rather than any particular speech or text database:
lexicons and grammars of various sorts are the most obvious examples.
The relevant aspects of these kinds of information also
often find natural expression in relational terms.

Users will commonly want to frame queries that combine information of
these kinds with predicates defined on annotation graphs: `find me all
the phrases flagged as questions produced by South Midland speakers
under the age of 30'.

The simplest way to permit this is simply to identify (some
of the) items in a relational database with (some of the) labels in an
annotation. This provides a limited, but useful, method for using the
results of certain relational queries in posing an annotational query,
or vice versa. More complex modes of interaction are also possible,
as are connections to other sorts of databases;
we regard this as a fruitful area for further research.

\subsubsection*{Generalizing time marks to an arbitrary ordering}

We have focused on the case of audio or video recordings, where a time
base is available as a natural way to anchor annotations. This role of
time can obviously be reassigned to any other well-ordered single
dimension.  The most obvious case is that of character- or
byte-offsets into an invariant text file. This is the principle used
in the so-called Tipster Architecture \cite{Grishman96},
where all annotations
are associated with stretches of an underlying text, identified via
byte offsets into a fixed file. We do not think that this method is
normally appropriate for use with audio transcriptions, because they
are so often subject to revision.

\subsubsection*{Generalizing node identifiers and arc labels}

As far as the annotation graph formalism is concerned, node
identifiers, arc types, and arc labels are just sets. As a
practical matter, members of each set would obviously be individuated
as strings. This opens the door to applications which encode
arbitrary information in these strings.
Indeed, the notion that arc labels encode `external' information is fundamental
to the enterprise.  The whole point of the annotations is to
include strings interpreted as orthographic words, speaker names,
phonetic segments, file references, or whatever. These interpretations
are not built into the formalism, however, and this is an equally
important trait, since it determines the simplicity and generality of
the framework.

In the current formalization, arcs are decorated with pairs consisting
of a type and a label.  This structure already contains a certain
amount of complexity, since the simplest kind of arc decoration would
be purely atomic.  In this case, we are convinced that the added value
provided by label types is well worth the cost: all the bodies of
annotation practice that we surveyed had some structure that was
naturally expressed in terms of atomic label types, and therefore a
framework in which arc decorations were just single uninterpreted
strings -- zeroth order labels -- would not be expressively adequate.

A first-order approach is to allow arcs to carry multiple
attributes and values -- what amounts to a fielded record.
The current formalization can be seen as providing records with just two fields.
It is easy to imagine a wealth of other possible fields.
Such fields could identify the original annotator and the creation
date of the arc.
They could represent the confidence level of some other field.
They could encode a complete history of successive modifications.
They could provide
hyperlinks to supporting material (e.g.\ chapter and verse in the
annotators' manual for a difficult decision).
They could provide equivalence class identifiers
(as a first-class part of the formalism rather than by
the external convention as in \S\ref{sec:multiple}).
And they could include an arbitrarily-long SGML-structured commentary.

In principle, we could go still further, and decorate arcs with
arbitrarily nested attribute-value matrices (AVMs) endowed with a type
system \cite{Carpenter92} -- a second-order approach.  These AVMs
could contain references to other parts of the annotation, and
multiple AVMs could contain shared substructures.  Substructures could
be disjoined to represent the existence of more than one choice, and
where separate choices are correlated the disjunctions could be
coindexed (i.e.\ parallel disjunction).  Appropriate attributes could
depend on the local type information.  A DTD-like label grammar could
specify available label types, their attributes and the type ordering
discussed in \S\ref{sec:hierarchy-local}.

We believe that this is a bad idea: it negates the effort that we made
to provide a simple formalism expressing the essential contents of
linguistic annotations in a natural and consistent way. Typed feature
structures are also very general and powerful devices, and entail
corresponding costs in algorithmic and implementational complexity.
Therefore, we wind up with a less useful representation that is much
harder to compute with.

Consider some of the effort that we have put into establishing a
simple and consistent ontology for annotation.  In the CHILDES case
(\S\ref{sec:childes}), we split a sentence-level annotation into a
string of word-level annotations for the sake of simplifying
word-level searches.  In the Festival case (\S\ref{sec:festival}) we
modeled hierarchical information using the syntactic chart
construction. Because of these choices, CHILDES and Festival
annotations become formally commensurate -- they can be searched or
displayed in exactly the same terms. With labels as typed feature
structures, whole sentences, whole tree structures, and indeed whole
databases could be packed into single labels. We could therefore have
chosen to translate CHILDES and Festival formats directly into typed
feature structures. If we had done this, however, the relationship
between simple concepts shared by the two formats -- such as lexical
tokens and time references -- would remain opaque.

For these reasons, we would like to remain cautious about adding to
the ontology of our formalism.  However, several simple extensions
seem well worth considering.  Perhaps the simplest one is to add
a single additional field to arc decorations, called the `comment',
which would be formally uninterpreted, but could be used in arbitrary
(and perhaps temporary) ways by implementations. It could be used to
add commentary, or to encode the authorship of the label, or indicate
who has permission to edit it, or in whatever other way. Another
possibility would be to add a field for encoding equivalence classes
of arcs directly, rather than by the indirect means specified earlier.

Our preference is to extend the formalism cautiously, where it
seems that many applications will want a particular capability, and
to offer a simple mechanism to permit local or experimental extensions,
while advising that it be used sparingly.

Finally, we note in passing that the same freedom for enriching arc
labels applies to node identifiers.  We have not given any examples in
which node identifiers are anything other than digit strings.
However, as with labels, in the general case a node identifier could
encode an arbitrarily complex data structure. For instance, it could be
used to encode the source of a time reference,
or to give a variant reference
(such as a video frame number, c.f.\ \S\ref{sec:offsets}),
or to specify whether a time reference is missing because it
is simply not known or it is inappropriate (c.f.\ \S\ref{sec:childes},
\ref{sec:lacito}).
Unlike the situation
with arc labels, this step is always harmless (except that
implementations that do not understand it will be left in the dark).
Only string identity matters to the formalism, and node identifiers do
not (in our work so far) have any standard interpretation outside the
formalism.

\subsection{Software}

We have claimed that annotation graphs can provide an interlingua for
varied annotation databases, a formal foundation for queries on such
databases, and a route to easier development and maintenance of such
databases. Delivering on these promises will require software.
Since we have made only some preliminary explorations so far, it would
be best to remain silent on the question until we have some
experience to report. However, for those readers who agree with us
that this is an essential point, we will sketch our current perspective.

As our catalogue of examples indicated, it is fairly easy to translate
between other speech database formats and annotation graphs, and we
have already built translators in several cases. We are also
experimenting with simple software for creation, visualization,
editing, validation, indexing, and search.  Our first goal
is an open collection of relatively
simple tools that are easy to prototype and to modify, in preference
to a monolithic `annotation graph environment.'  However, we are
also committed to the idea that tools for creating and using
linguistic annotations should be widely accessible to computationally
unsophisticated users, which implies that eventually such tools need
to be encapsulated in reliable and simple interactive form.

Other researchers have also begun to experiment with the annotation
graph concept as a basis for their software tools, and a key index
of the idea's success will of course be the extent to which tools are
provided by others.

\subsubsection*{Visualization, creation, editing}

Existing open-source software such as Transcriber \cite{Barras98},
Snack \cite{Sjoelander98},
and the ISIP transcriber tool
[\smtt{www.} \smtt{isip.msstate.edu/resources/software}],
whose user interfaces are all implemented in Tcl/tk,
make it easy to create interactive tools for creation, visualization,
and editing of annotation graphs.

For instance, Transcriber can be used without any changes to produce
transcriptions in the LDC Broadcast News format, which can then be translated
into annotation graphs. Provision of simple input/output functions enables the
program to read and write annotation graphs directly. The architecture
of the current tool is not capable of dealing with arbitrary
annotation graphs, but generalizations in that direction are planned.

\subsubsection*{Validation}

An annotation may need to be submitted to a variety of validation
checks, for basic syntax, content and larger-scale structure.

First, we need to be able to tokenize and parse an annotation, without
having to write new tokenizers and parsers for each new task.
We also need to undertake some superficial syntax checking,
to make sure that brackets and quotes balance, and so on.
In the SGML realm, this need is partially met by DTDs.
We propose to meet the same need by developing
conversion and creation tools
that read and write well-formed graphs, and by input/output modules that
can be used in the further forms of validation cited below.

Second, various content checks need to be performed.  For instance,
are purported phonetic segment labels actually members of a designated
class of phonetic symbols or strings?  Are things marked as
`non-lexemic vocalizations' drawn from the officially approved list?
Do regular words appear in the spell-check dictionary?  Do capital letters
occur in legal positions?  These checks are not difficult to
implement,  e.g.\ as Perl scripts, especially given a module for
handling basic operations correctly.

Finally, we need to check for correctness of hierarchies of arcs.
Are phonetic segments all inside words, which are all inside phrases,
which are all inside conversational turns, which are all inside conversations?
Again, it is easy to define such checks in a software environment
that has appropriately expressive primitives
(e.g.\ a Perl annotation graph module).

\subsubsection*{Indexing and Search}

Indexing of the types discussed earlier (\S\ref{sec:indexing}), is
well defined, algorithmically simple, and easy to implement in a
general way. Construction of general query systems, however,
is a matter that needs to be explored more fully in order to decide
on the details of the query primitives and the methods for building
complex queries, and also to try out different ways to express
queries. Among the many questions to be explored are:
\begin{enumerate}
\item how to express general graph- and time-relations;
\item how to integrate regular expression matching over labels;
\item how to integrate annotation-graph queries and relational queries;
\item how to integrate lexicons and other external resources;
\item how to model sets of databases, each of which
contains sets of annotation graphs, signals and perhaps relational
side-information.
\end{enumerate}

It is easy to come up with answers to each of these questions,
and it is also easy to try the answers out, for instance in the context of
a collection of Perl modules providing the needed primitive operations.
We regard it as an open research problem to find good answers
that interact well, and also to find good ways to express queries in the
system that those answers will define.

\subsection{Envoi}


Whether or not our ideas are accepted by the various research
communities who create and use linguistic annotations, we hope to
foster discussion and cooperation among members of these communities.
A focal point of this effort is the Linguistic Annotation Page at
[\smtt{www.ldc.upenn.edu/annotation}].  

When we look at the numerous and diverse forms of linguistic
annotation documented on that page, we see underlying similarities
that have led us to imagine general methods for access and search, and
shared tools for creation and maintenance. We hope that this discussion
will move others in the same direction.

\section{Acknowledgements}

An earlier version of this paper was presented at ICSLP-98.
We are grateful to the following people for
discussions which have helped clarify our
ideas about annotations, and for comments on earlier drafts:
Peter Buneman,
Steve Cassidy,
Chris Cieri,
Hamish Cunningham,
David Graff,
Ewan Klein,
Brian MacWhinney,
Boyd Michailovsky,
Florian Schiel,
Richard Sproat,
Paul Taylor,
Henry Thompson,
Peter Wittenburg,
Jonathan Wright,
and participants of the COCOSDA workshop at ICSLP-98.

\vfil\pagebreak
\raggedright


\begin{thebibliography}{10}

\bibitem{Altosaar98}
T.~Altosaar, M.~Karjalainen, M.~Vainio, and E.~Meister.
\newblock {Finnish} and {Estonian} speech applications developed on an
  object-oriented speech processing and database system.
\newblock In {\em Proceedings of the First International Conference on Language
  Resources and Evaluation Workshop: Speech Database Development for Central
  and Eastern European Languages}, 1998.
\newblock Granada, Spain, May 1998.

\bibitem{Anderson91}
A.~Anderson, M.~Bader, E.~Bard, E.~Boyle, G.~M. Doherty, S.~Garrod, S.~Isard,
  J.~Kowtko, J.~McAllister, J.~Miller, C.~Sotillo, H.~Thompson, and R.~Weinert.
\newblock The {HCRC} {Map} {Task} corpus.
\newblock {\em Language and Speech}, 34:351--66, 1991.

\bibitem{Barras98}
Claude Barras, Edouard Geoffrois, Zhibiao Wu, and Mark Liberman.
\newblock Transcriber: a free tool for segmenting, labelling and transcribing
  speech.
\newblock In {\em Proceedings of the First International Conference on Language
  Resources and Evaluation}, 1998.

\bibitem{Bird95}
Steven Bird.
\newblock {\em Computational Phonology: A Constraint-Based Approach}.
\newblock Studies in Natural Language Processing. Cambridge University Press,
  1995.

\bibitem{Bird97sigphon}
Steven Bird.
\newblock A lexical database tool for quantitative phonological research.
\newblock In {\em Proceedings of the Third Meeting of the ACL Special Interest
  Group in Computational Phonology}. Association for Computational Linguistics,
  1997.

\bibitem{BirdKlein90}
Steven Bird and Ewan Klein.
\newblock Phonological events.
\newblock {\em Journal of Linguistics}, 26:33--56, 1990.

\bibitem{BirdLadd91}
Steven Bird and D.~Robert Ladd.
\newblock Presenting autosegmental phonology.
\newblock {\em Journal of Linguistics}, 27:193--210, 1991.

\bibitem{Browman89}
Catherine Browman and Louis Goldstein.
\newblock Articulatory gestures as phonological units.
\newblock {\em Phonology}, 6:201--51, 1989.

\bibitem{Brown90}
Peter~F. Brown, John Cocke, Stephen~A. {Della Pietra}, Vincent~J. {Della
  Pietra}, Fredrick Jelinek, Robert~L. Mercer, and Paul~S. Roossin.
\newblock A statistical approach to machine translation.
\newblock {\em Computational Linguistics}, 16:79--85, 1990.

\bibitem{Carpenter92}
Bob Carpenter.
\newblock {\em The Logic of Typed Feature Structures}, volume~32 of {\em
  Cambridge Tracts in Theoretical Computer Science}.
\newblock Cambridge University Press, 1992.

\bibitem{Cassidy96}
Steve Cassidy and Jonathan Harrington.
\newblock Emu: An enhanced hierarchical speech data management system.
\newblock In {\em Proceedings of the Sixth Australian International Conference
  on Speech Science and Technology}, 1996.
\newblock [www.shlrc.mq.edu.au/emu/].

\bibitem{Battista94}
Giuseppe {Di Battista}, Peter Eades, Roberto Tamassia, and Ioannis~G. Tollis.
\newblock Algorithms for drawing graphs: an annotated bibliography.
\newblock [wilma.cs.brown/edu/pub/papers/compgeo/gdbiblio.ps.gz], 1994.

\bibitem{MATE-D1.2}
Laila Dybkj{\ae}r, Niels~Ole Bernsen, Hans Dybkj{\ae}r, David McKelvie, and
  Andreas Mengel.
\newblock The mate markup framework.
\newblock MATE Deliverable D1.2, Odense University, 1998.

\bibitem{Ehlich92}
Konrad Ehlich.
\newblock {HIAT} -- a transcription system for discourse data.
\newblock In Jane~A. Edwards and Martin~D. Lampert, editors, {\em Talking Data:
  Transcription and Coding in Discourse Research}, pages 123--48. Hillsdale,
  NJ: Erlbaum, 1992.

\bibitem{TIMIT86}
John~S. Garofolo, Lori~F. Lamel, William~M. Fisher, Jonathon~G. Fiscus,
  David~S. Pallett, and Nancy~L. Dahlgren.
\newblock {\em The {DARPA TIMIT} Acoustic-Phonetic Continuous Speech Corpus
  {CDROM}}.
\newblock NIST, 1986.
\newblock [www.ldc.upenn.edu/lol/docs/TIMIT.html].

\bibitem{GazdarMellish89}
Gerald Gazdar and Chris Mellish.
\newblock {\em Natural Language Processing in Prolog: An Introduction to
  Computational Linguistics}.
\newblock Addison-Wesley, 1989.

\bibitem{Godfrey92}
J.~J. Godfrey, E.~C. Holliman, and J.~McDaniel.
\newblock Switchboard: A telephone speech corpus for research and develpment.
\newblock In {\em Proceedings of the IEEE Conference on Acoustics, Speech and
  Signal Processing}, volume~I, pages 517--20, 1992.

\bibitem{Greenberg96}
S.~Greenberg.
\newblock The switchboard transcription project.
\newblock LVCSR Summer Research Workshop, Johns Hopkins University, 1996.

\bibitem{Grishman96}
R.~Grishman.
\newblock {TIPSTER Architecture Design Document Version 2.3}.
\newblock Technical report, DARPA, 1997.
\newblock [www.nist.gov/itl/div894/894.02/related\_projects/tipster/].

\bibitem{Harrington93}
Jonathan Harrington, Steve Cassidy, Janet Fletcher, and A.~McVeigh.
\newblock The {Mu+} speech database system.
\newblock {\em Computer Speech and Language}, 7:305--31, 1993.

\bibitem{Hertz90}
Susan~R. Hertz.
\newblock The delta programming language: an integrated approach to nonlinear
  phonology, phonetics, and speech synthesis.
\newblock In John Kingston and Mary~E. Beckman, editors, {\em Papers in
  Laboratory Phonology I: Between the Grammar and Physics of Speech},
  chapter~13, pages 215--57. Cambridge University Press, 1990.

\bibitem{JuretafskyBates97}
Daniel Jurafsky, Rebecca Bates, Noah Coccaro, Rachel Martin, Marie Meteer,
  Klaus Ries, Elizabeth Shriberg, Andreas Stolcke, Paul Taylor, and Carol {Van
  Ess-Dykema}.
\newblock Automatic detection of discourse structure for speech recognition and
  understanding.
\newblock In {\em Proceedings of the 1997 IEEE Workshop on Speech Recognition
  and Understanding}, pages 88--95, Santa Barbara, 1997.

\bibitem{JurafskyShriberg97}
Daniel Jurafsky, Elizabeth Shriberg, and Debra Biasca.
\newblock {S}witchboard {SWBD-DAMSL} {L}abeling {P}roject {C}oder's {M}anual,
  {D}raft 13.
\newblock Technical Report 97-02, University of Colorado Institute of Cognitive
  Science, 1997.
\newblock [stripe.colorado.edu/\~{}jurafsky/manual.august1.html].

\bibitem{MacWhinney95}
Brian MacWhinney.
\newblock {\em The CHILDES Project: Tools for Analyzing Talk}.
\newblock Mahwah, NJ: Lawrence Erlbaum., second edition, 1995.
\newblock [poppy.psy.cmu.edu/childes/].

\bibitem{Marcus93}
Mitchell~P. Marcus, Beatrice Santorini, and Mary~Ann Marcinkiewicz.
\newblock Building a large annotated corpus of {English}: The {Penn}
  {Treebank}.
\newblock {\em Computational Linguistics}, 19(2):313--30, 1993.
\newblock www.cis.upenn.edu/~treebank/home.html.

\bibitem{Michailovsky98}
Boyd Michailovsky, John~B. Lowe, and Michel Jacobson.
\newblock Linguistic data archiving project.
\newblock [lacito.vjf.cnrs.fr/ARCHIVAG/ENGLISH.htm].

\bibitem{Neidle98}
Carol Neidle and D.~MacLaughlin.
\newblock {SignStream$^{\mbox{\scriptsize TM}}$}: a tool for linguistic
  research on signed languages.
\newblock {\em Sign Language and Linguistics}, 1:111--14, 1998.
\newblock [web.bu.edu/asllrp/SignStream].

\bibitem{UTF98}
NIST.
\newblock A universal transcription format {(UTF)} annotation specification for
  evaluation of spoken language technology corpora.
\newblock [www.nist.gov/speech/hub4\_98/utf-1.0-v2.ps], 1998.

\bibitem{Sacks-Davis97}
Ron Sacks-Davis, Tuong Dao, James~A. Thom, and Justin Zobel.
\newblock Indexing documents for queries on structure, content and attributes.
\newblock In {\em International Symposium on Digital Media Information Base},
  pages 236--45, 1997.

\bibitem{Schegloff98}
Emanuel Schegloff.
\newblock Reflections on studying prosody in talk-in-interaction.
\newblock {\em Language and Speech}, 41:235--60, 1998.
\newblock www.sscnet.ucla.edu/soc/faculty/schegloff/prosody/.

\bibitem{Schiel98}
Florian Schiel, Susanne Burger, Anja Geumann, and Karl Weilhammer.
\newblock The {Partitur} format at {BAS}.
\newblock In {\em Proceedings of the First International Conference on Language
  Resources and Evaluation}, 1998.
\newblock [www.phonetik.uni-muenchen.de/Bas/BasFormatseng.html].

\bibitem{Sjoelander98}
K{\aa}re Sj\"{o}lander, Jonas Beskow, Joakim Gustafson, Erland Lewin, Rolf
  Carlson, and Bj\"orn Granstr\"om.
\newblock Web-based educational tools for speech technology.
\newblock In {\em ICSLP-98}, 1998.

\bibitem{Taylor95}
Ann Taylor.
\newblock {\em Dysfluency Annotation Stylebook for the Switchboard Corpus}.
\newblock University of Pennsylvania, Department of Computer and Information
  Science, 1995.
\newblock [ftp.cis.upenn.edu/pub/treebank/swbd/doc/DFL-book.ps].

\bibitem{Taylor98}
Paul~A. Taylor, Alan~W. Black, and Richard~J. Caley.
\newblock The architecture of the {Festival} speech synthesis system.
\newblock In {\em Third International Workshop on Speech Synthesis}, Sydney,
  Australia, November 1998.

\bibitem{Taylor99}
Paul~A. Taylor, Alan~W. Black, and Richard~J. Caley.
\newblock Heterogeneous relation graphs as a mechanism for representing
  linguistic information.
\newblock [www.cstr.ed.ac.uk/publications/new/draft/Taylor\_draft\_a.ps], 1999.

\bibitem{TEI-P3}
{Text Encoding Initiative}.
\newblock {\em Guidelines for Electronic Text Encoding and Interchange (TEI
  P3)}.
\newblock Oxford University Computing Services, 1994.
\newblock [www.uic.edu/orgs/tei/].

\bibitem{Thompson97}
Henry~S. Thompson and David McKelvie.
\newblock Hyperlink semantics for standoff markup of read-only documents.
\newblock In {\em SGML Europe '97}, 1997.
\newblock [www.ltg.ed.ac.uk/\~{}ht/sgmleu97.html].

\end{thebibliography}
\end{document}